%% file: main.tex
\documentclass[10pt,twocolumn,letterpaper]{article}

\usepackage{cvpr}
\usepackage{times}
\usepackage{epsfig}
\usepackage{graphicx}
\usepackage{amsmath}
\usepackage{amssymb}
\usepackage{soul}

\newcommand{\comment}[1]{}

\DeclareMathOperator*{\argmin}{arg\,min}

\newcommand{\matr}[1]{\mathbf{#1}}

\renewcommand{\paragraph}[1]{\vspace{2mm} \noindent \textbf{#1}}

\newcounter{alphasect}
\def\alphainsection{0}

\let\oldsection=\subsection
\def\subsection{%
  \ifnum\alphainsection=1%
    \addtocounter{alphasect}{1}
  \fi%
\oldsection}%

\renewcommand\thesubsection{%
  \ifnum\alphainsection=1%
    \Alph{alphasect}
  \else%
    \arabic{section}.\arabic{subsection}
  \fi%
}%

\newenvironment{alphasection}{%
  \ifnum\alphainsection=1%
    \errhelp={Let other blocks end at the beginning of the next block.}
    \errmessage{Nested Alpha section not allowed}
  \fi%
  \setcounter{alphasect}{0}
  \def\alphainsection{1}
}{%
  \setcounter{alphasect}{0}
  \def\alphainsection{0}
}%

\usepackage[pagebackref=true,breaklinks=true,letterpaper=true,colorlinks,bookmarks=false]{hyperref}

\cvprfinalcopy 

\def\cvprPaperID{9046} 

\ifcvprfinal\fi
\begin{document}

\title{Deep Geometric Functional Maps: Robust \\Feature Learning for Shape Correspondence}

\author{Nicolas Donati\\
LIX, \'Ecole Polytechnique\\
{\tt\small nicolas.donati@polytechnique.edu}
\and
Abhishek Sharma\\
LIX, \'Ecole Polytechnique\\
{\tt\small kein.iitian@gmail.com}
\and
Maks Ovsjanikov\\
LIX, \'Ecole Polytechnique\\
{\tt\small maks@lix.polytechnique.fr}
}

\maketitle

\begin{abstract}
  We present a novel learning-based approach for computing correspondences between non-rigid 3D shapes. Unlike previous methods that either require extensive training data or operate on handcrafted input descriptors and thus generalize poorly across diverse datasets, our approach is both accurate and robust to changes in shape structure. Key to our method is a feature-extraction network that learns directly from raw shape geometry, combined with a novel regularized map extraction layer and loss, based on the functional map representation. We demonstrate through extensive experiments in challenging shape matching scenarios that our method can learn from less training data than existing supervised approaches and generalizes significantly better than current descriptor-based learning methods. Our source code is available at: \url{https://github.com/LIX-shape-analysis/GeomFmaps}.
\end{abstract}


\input{sections/introduction.tex}

\input{sections/related_work.tex}

\input{sections/background.tex}
\label{sec:background}
\input{sections/method.tex}

\input{sections/results.tex}

\input{sections/conclusion.tex}

{\small
\bibliographystyle{ieee_fullname}
\bibliography{egbib.bib}
}


\section{Supplement}
\begin{alphasection}
\input{sections/supplementary_material.tex}

\end{alphasection}

\end{document}

%% file: sections/introduction.tex
\section{Introduction}
\label{sec:intro}

\begin{figure}[t]
\begin{center}
   \includegraphics[width=0.9\linewidth]{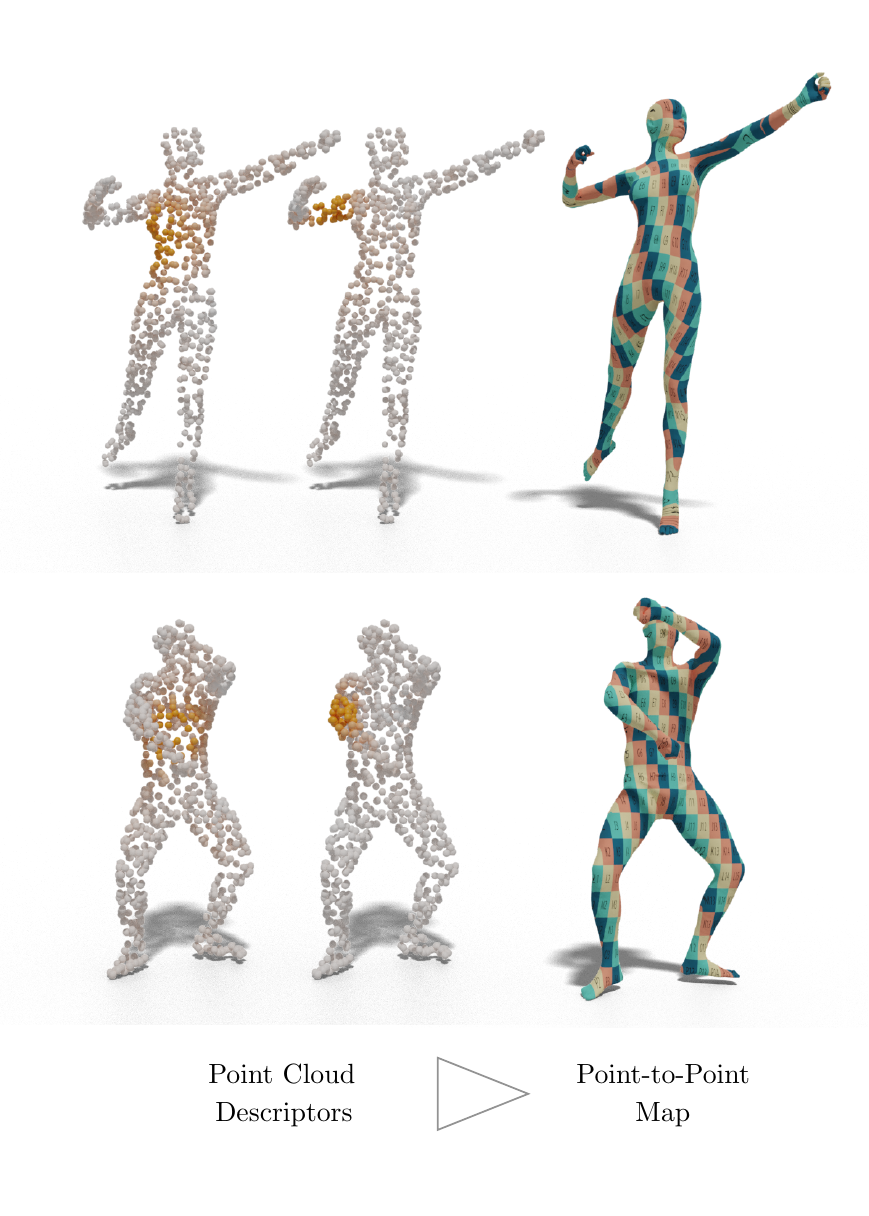}
\end{center}
\vspace{-5mm}
   \caption{Given a pair of shapes, our approach builds consistent descriptors directly from the underlying point clouds (left), and automatically computes an accurate pointwise correspondence (right).
\label{fig:teaser_main}
\vspace{-5mm}
}
\end{figure}

Shape correspondence is a key problem in computer vision, computer graphics and related fields with a broad range of applications, including texture or deformation transfer and statistical shape analysis \cite{bogo2014}, among many others. While classical correspondence methods have been based on handcrafted features or deformation models \cite{van2011survey}, more recent approaches have focused on \emph{learning} an optimal model from the data either in supervised \cite{corman2014supervised,wei2016dense,litany2017deep,groueix20183d} or even unsupervised settings \cite{halimi2018self,roufosse2019unsupervised,groueix2019unsupervised}. 

Despite significant progress in recent years, however, learning-based approaches for shape correspondence typically require large amounts of training data in order to learn a model that generalizes well to diverse shape classes \cite{wei2016dense,groueix20183d}. Several existing methods address this challenge by learning a derived representation, through a non-linear transformation of pre-computed feature descriptors \cite{corman2014supervised,litany2017deep,halimi2018self,roufosse2019unsupervised}, rather than on the geometry of the shapes themselves. Unfortunately, as we demonstrate below, this reliance on \emph{a priori} hand-crafted descriptors makes the resulting learned models both less robust and less accurate leading to a significant drop in generalization power to new shape classes or instances. 

In this work, we propose an approach that combines the power of learning directly from the 3D shapes with strong regularization based on a novel spectral correspondence extraction layer. Our method is inspired by recent learning techniques employing the functional map representation \cite{litany2017deep,roufosse2019unsupervised}; however, we extend them to learn the features from 3D geometry rather than from some pre-computed descriptors. Furthermore, we introduce a regularizer into the functional map computation layer that greatly improves the speed and robustness of training. Finally, we demonstrate how the spectral loss based on the functional map representation in the reduced basis significantly reduces over-fitting, while still leading to accurate correspondences coupled with recent post-processing techniques. As a result, our overall pipeline is both more robust and has greater generalization power than existing methods, while still being able to learn from limited training data. 

%% file: sections/related_work.tex
\section{Related Work}
\label{sec:related}

Computing point-to-point maps between two 3D discrete surfaces is a very well-studied area of computer vision. Below, we review those closest to our method, or with the best known results to serve as baselines, and refer to recent surveys \cite{van2011survey,biasotti2016recent,sahilliouglu2019recent}  for an in-depth discussion.

Our method is built upon the functional map representation, which was originally introduced in \cite{ovsjanikov2012functional} as a tool for non-rigid shape matching, and then extended in follow-up works \cite{ovsjanikov2017course}. 
The key property of this representation is being able to express maps as small matrices, encoded in a reduced basis, which greatly simplifies the associated optimization problems.

The original work used only a basic set of constraints on functional maps, which have been extended significantly in, e.g., \cite{kovnatsky2013coupled,aflalo2013spectral,huang2014functional,eynard2016coupled,burghard2017embedding,rodola2017partial,commutativity,huang2017adjoint,ren2018continuous} among many other works. These approaches both extend the generality and improve the robustness of the functional map estimation pipeline, by using regularizers, robust penalties and powerful post-processing of the computed maps.

A key challenge in all of functional map estimation techniques, however, is the strong reliance on given input \emph{descriptors} used for computing the maps. Several approaches have suggested to use robust norms \cite{kovnatsky2013coupled,kovnatsky2015functional}, improved pointwise map recovery \cite{rodola-vmv15,ezuz2017deblurring} or more principled regularizers \cite{ren2019structured} which can help alleviate noise in the input descriptors to a certain extent but do not resolve strong inconsistencies in challenging cases.

More recent techniques have advocated learning optimal descriptors for functional map estimation directly from the data  \cite{corman2014supervised,litany2017deep}. These methods compute a transformation of given input descriptors so that the estimated functional maps are close to ground truth maps given during training. This idea was very recently extended to the \emph{unsupervised} setting \cite{halimi2018self,roufosse2019unsupervised} where the supervised loss was replaced with structural penalties on the computed maps.

Despite significant progress, however, in all of these cases, the descriptors are optimized through a transformation of \emph{hand-crafted input features}, such as SHOT \cite{shot}, Heat \cite{hks} or Wave kernel signatures \cite{wks}. This has two severe consequences: first, any information not present in the input features will be absent from the optimized descriptors, and second, such approaches generalize poorly across datasets as the input features can change significantly. This is particularly true of the commonly-used SHOT descriptors \cite{shot}, which are sensitive to the triangle mesh structure and, as we show below, can vary drastically across different datasets.


A number of other techniques have also been proposed for shape correspondence learning without using the functional map representation. These include approaches that exploit novel convolutional layers on triangle meshes \cite{masci2015geodesic,MasBosBroVan16,monti2017,poulenard2018topological} and more general methods that use learning from depth-maps \cite{wei2016dense} or in some feature space \cite{sun2017deep,chen2018jointly} among many others. Remarkably, relatively few methods aim to learn directly from the raw 3D shape geometry for shape correspondence, with the notable exceptions of \cite{groueix20183d,deprelle2019learning}. In large part this is due to the complexity of the correspondence problem, where unlike, e.g., shape segmentation, the number of labels can be unbounded. As a result, existing techniques address this either by learning from precomputed features, or relying on template-based matching and large training sets \cite{groueix20183d,deprelle2019learning}, that might even require manual curation. Although PointNet \cite{qi2017pointnet}
and its variants \cite{2017Qipointnet2, pcnn_extension_operator,thomas2019KPConv} achieve impressive results from raw point clouds for classification tasks, they are not yet competitive for shape correspondence task. 

\subsection*{Contribution}
In this paper we show that feature learning for shape matching can be done directly from the raw 3D geometry even in the presence of relatively little training data, and without relying on a template or an \emph{a priori} parametric (e.g., human body) model. Our main contribution is a end-to-end learnable pipeline that computes features from the 3D shapes and uses them for accurate dense point-to-point correspondence. We achieve this by introducing a novel map extraction layer using the functional map representation in a reduced basis, which provides a very strong regularization. Finally, we demonstrate that recent refinement techniques adapted to small functional maps \cite{melzi2019zoomout}, combined with our efficient learning pipeline jointly result in accurate dense maps at the fraction of the cost of existing methods.

%% file: sections/background.tex
\section{Shape Matching and Functional Maps}

\label{subsec:fmaps}
One of the building blocks in our pipeline work is based on the functional map framework and representation. For completeness, we briefly review the basic
notions for estimating functional maps, and refer the interested reader to a recent course
\cite{ovsjanikov2017course} for a more in-depth discussion.


\paragraph{Basic Pipeline} Given a pair of 3D shapes, $\mathcal{M}, \mathcal{N}$ represented in a discrete setting as triangle meshes, and containing respectively $m$ and $n$ vertices, this pipeline aims at computing a map between them.




It consists in four main steps. First, the first few eigenfunctions of the discrete Laplace-Beltrami operator are computed on  each shape, namely $k_{\mathcal{M}}$ and $k_{\mathcal{N}}$ functions respectively.

Second, a set of descriptor \emph{functions} on each shape that are expected to be approximately preserved by the unknown map. For instance, a descriptor function can correspond to a particular dimension of the Heat or Wave Kernel Signatures \cite{hks,wks} computed at each point. Their coefficients are stored in the respective basis as columns in matrices $\matr{A}, \matr{B}$.

Third, the optimal \emph{functional map} $\matr{C}$ is then computed by solving the following optimization problem:
\begin{align}
\label{eq:opt_problem}
\matr{C_{\text{opt}}} = \argmin_{\matr{C}} E_{\text{desc}}\big(\matr{C}\big) + \alpha E_{\text{reg}}\big(\matr{C}\big),
\end{align}
where the first term aims at preserving the descriptors: $E_{\text{desc}}\big(\matr{C}\big)
= \big\Vert \matr{C} \matr{A} - \matr{B}\big\Vert^2$, whereas the second term regularizes the map by promoting the correctness of its overall structural properties. It is common to use Fröbenius norm to compute the distance between these matrices. This Eq.~(\ref{eq:opt_problem}) leads to a simple least-squares problem with  $k_{\mathcal{M}} \times k_{\mathcal{N}}$ unknowns, independent of the number of points on the shapes.

         
As a last step, the estimated functional map $\matr{C}$, which maps across the spectral domains and converted to a point-to-point map. As a post processing step, called refinement, a number of advanced techniques are available \cite{rodola-vmv15,ezuz2017deblurring, ren2018continuous, melzi2019zoomout}. Most of them iteratively take the map from spectral to spatial domain, until it reaches a local optimum.

\subsection{Deep Functional Maps}

Despite its simplicity and efficiency, being a sequential framework, the functional map estimation pipeline described above is fundamentally error prone, due to the initial choice of descriptor functions. To alleviate this dependence, several approaches have been proposed to learn an optimal transformation of initial descriptors from data \cite{corman2014supervised,litany2017deep,roufosse2019unsupervised}. These works aim at transforming a given set of descriptors so that the optimal computed map satisfies some desired criteria during training. This transformation can be learned with a supervised loss, as in \cite{corman2014supervised,litany2017deep}, as well as with an unsupervised loss as in the more recent works of \cite{halimi2018self,roufosse2019unsupervised}.

More specifically, the FMNet approach proposed in \cite{litany2017deep} assumes to have as input, a set of shape pairs for which ground truth point-wise maps are known, and aims to solve the following problem:
\begin{align}
\label{eq:fmnet1}
&\min_{T} \sum_{(S_1,S_2) \in \text{Train}} l_F (Soft(\mathbf{C}_{\text{opt}}), GT_{(S_1, S_2)}), \text{ where }\\
\label{eq:fmnet2}
&\matr{C}_{\text{opt}} = \argmin_{\matr{C}} \| \matr{C} \matr{A}_{T(D_1)} - \matr{A}_{T(D_2)} \|.
\end{align}
Here, adopting the notation from \cite{roufosse2019unsupervised} $T$ is a non-linear transformation, in the form of a neural network, to be applied to some input descriptor functions $D$, $\text{Train}$ is the set of training pairs for which ground truth correspondence $GT_{(S_1, S_2)}$ is known, $l_F$ is the \emph{soft error loss}, which penalizes the deviation of the computed functional map $\mathbf{C}_{\text{opt}}$, after converting it to a soft map $Soft(\mathbf{C}_{\text{opt}})$ from the ground truth correspondence, and $\mathbf{A}_{T(D_1)}$ denotes the transformed descriptors $D_1$ written in the basis of shape $1$. Thus, the FMNet framework \cite{litany2017deep} learns a transformation $T$ of descriptors $T(D_1)$, $T(D_2)$ based on a supervised loss that minimizes the discrepancy between the resulting soft map and the known ground truth correspondence.

A related recent approach, SURFMNet \cite{roufosse2019unsupervised} follows a similar strategy but replaces $l_{F}$ with an \emph{unsupervised loss} that enforces the desired structural properties on the resulting map, such as its bijectivity, orthonormality and commutativity with the Laplacian. 

\paragraph{3D-CODED}
In contrast to the the methods described above that primarily operate in the \emph{spectral} domain, there are also some approaches that never leave the spatial domain. With the recent works on point clouds neural networks, pioneered by PointNet \cite{qi2017pointnet}, and significantly extended by \cite{pcnn_extension_operator, thomas2019KPConv}, to name a few, it is now possible to learn 3D features directly from point clouds. 3D-CODED  \cite{groueix20183d, deprelle2019learning} is based on this approach, as it is a method built on a variational auto-encoder with a PointNet architecture for the encoder. Their method relies on a template that is supposed to be deformable in a non-rigid but isometric way to any of the shape of the datasets. It is a supervised method, and requires the knowledge of all ground-truth correspondences between any shape of the dataset and the deformable template.
3D-CODED is trained on 230K shapes, introduced in SURREAL \cite{varol17_surreal}, and generated with SMPL \cite{SMPL_2015}.


\paragraph{Motivation}
The two main classes of existing approaches have their associated benefits and drawbacks. On the one hand, spectral methods are able to use small matrices instead of all the points of the shape, and operate on \emph{intrinsic} properties of the 3D surfaces, making them resilient to a change in pose, and allowing them to train on really small datasets. However, due to their use of input descriptors (typically SHOT \cite{shot}), they tend to overfit to the connectivity of the training set, which can lead to catastrophic results even in apparently simple cases. On the other hand, 3D-CODED shows extreme efficiency when trained on enough data, regardless of the connectivity, but with a small dataset, it is prone to overfitting and fails to generalize the training poses  to predict the different poses of the test set.

Our method is a mix of the two approaches, and, as we show below, can obtain accurate results with little training data leading to state-of-the-art accuracy on a challenging recent benchmark of human shapes in different poses and with different connectivity \cite{SHREC19}.

%% file: sections/method.tex
\section{Method}
\label{sec:method}


\subsection{Overview}

In this paper, we introduce a novel approach to learning descriptors on shapes in order to get correspondences through the functional map framework. Our method is composed of two main parts, labeled as \textit{Feat} and \textit{FMReg} in Figure \ref{fig:architecture}. The first aims at optimizing point cloud convolutional filters \cite{pcnn_extension_operator, thomas2019KPConv} to extract features from the raw geometry of the shapes. These filters are learned using a Siamese network on the source and a target shapes by using shared learnable parameters $\Theta$, in a similar way as in \cite{litany2017deep}. However, unlike that approach and follow-up works \cite{halimi2018self,roufosse2019unsupervised} we learn the features directly from the geometry of the shapes rather than computing a transformation of some pre-defined existing descriptors. These learned descriptors are projected in the spectral bases of the shapes and fed to the second block of the method, which uses them in a novel regularized functional map estimation layer. Finally, we use a spectral loss, based on the difference between the computed and the ground truth functional maps. This makes our approach very efficient as it operates purely in the spectral domain, avoiding expensive geodesic distance matrix computations as in \cite{litany2017deep,halimi2018self} and moreover allows us to handle functional or soft ground truth input maps without requiring the training shapes to have the same number of points or fixed mesh connectivity. 

We stress again that the two components: learning features directly from the shapes and using the functional map representation both play a crucial role in our setup. The former allows us to learn robust and informative features independently from the mesh structure, while the latter allows us to strongly regularize correspondence learning, resulting in a method that generalizes even in the presence of a relatively small training set.


\subsection{Architecture}

\begin{figure}[t]
\begin{center}
   \includegraphics[width=0.95\linewidth]{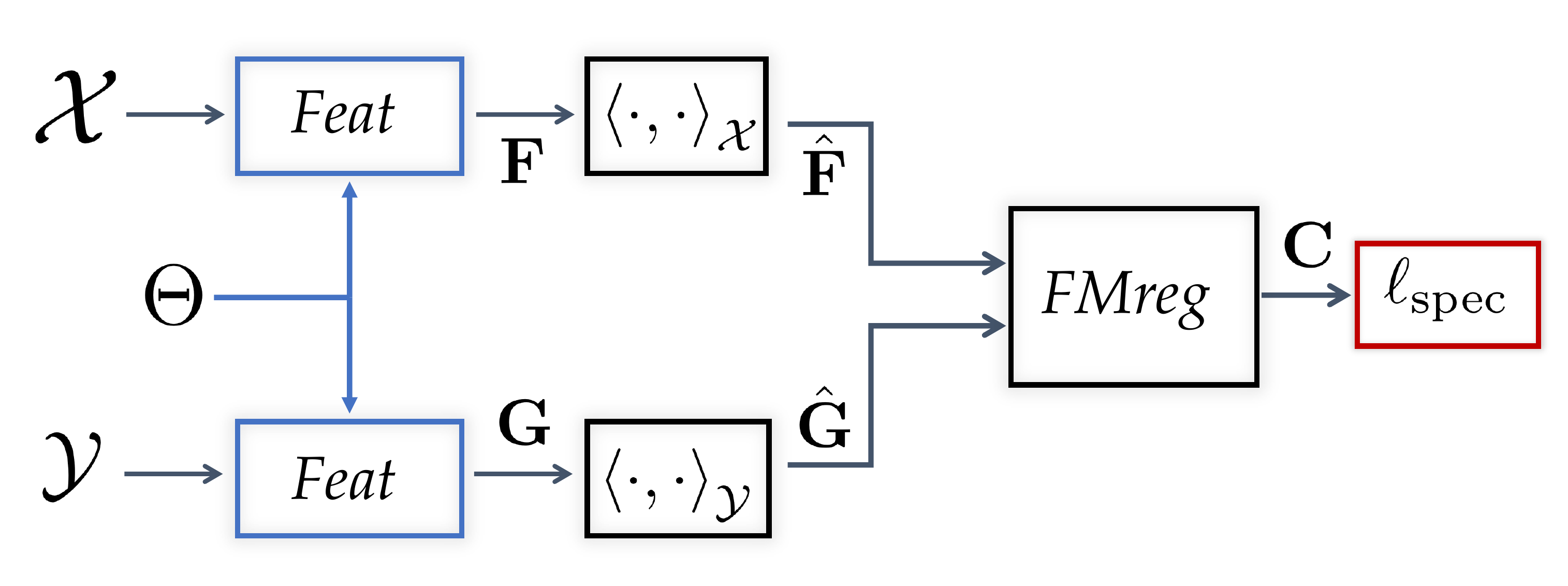}
\end{center}
   \caption{Overview of our approach: given a pair of shapes, we optimize for a point cloud convolutional model to get point-wise features for each shape, that we convert to a functional map using our FMReg block. The loss that we put forward penalizes maps according to their distance to the ground-truth map between the two shapes.}
\label{fig:architecture}
\end{figure}

The novelty of our architecture lies in its hybrid character. The first part, which we will refer to as the \textit{feature extractor} in the following, aims at computing point-wise features on the input shapes. It corresponds to the \textit{Feat} block in Figure \ref{fig:architecture}, and takes as input only the point clouds making it robust towards changes in connectivity.

The purpose of the second part is to recover robust functional maps using these learned features. This block is built according to the pipeline of \cite{ovsjanikov2012functional}, first taking the features to the spectral domain over the two shapes (which corresponds to the dot products blocks after the \emph{Feat} blocks in Figure \ref{fig:architecture}), and then computing the map by minimizing an energy. However, since our method is based on a neural network, this operation should be differentiable with respect to the features over the shapes for the back-propagation algorithm to work. We extend the previously proposed functional map layers \cite{litany2017deep} to also incorporate a differentiable regularizer, which results in the very robust map extraction, represented as \textit{FMReg} in Figure \ref{fig:architecture}.

\subsection{The feature extractor}
\label{subsec:archi_feat}

The goal of this block is to learn functional characterizations of point clouds that will later be used to compute spectral descriptors and then functional maps. To this end, this network must be applied with the same weights to the source and target shapes, as represented in Figure \ref{fig:architecture}, and must result in informative descriptors, extracted from the point clouds of the two shapes.

For this part, we chose the state of the art point cloud learning method KPConv \cite{thomas2019KPConv}, by extending the segmentation network proposed in that work. Our feature extractor is thus a Siamese version of the segmentation network described in KPConv, which we review for completeness in the supplementary materials.

\subsection{The regularized FMap layer}

This block provides a novel fully differentiable way to compute a robust functional map from potentially low dimensional spectral descriptors.


The main goal is, as in Section \ref{subsec:fmaps}, to recover the ground-truth bijection between $\mathcal{M}$ and $\mathcal{N}$, on which we have the computed raw-data features $\mathbf{F}$ and $\mathbf{G}$.

For this we first express the computed feature functions in the respective spectral basis, which we denote by $\Phi^{\mathcal{M}}$ and $\Phi^{\mathcal{N}}$. This leads to the spectral descriptors $\mathbf{A} = (\Phi^{\mathcal{M}})^{\dagger} \mathbf{F}$ and $\mathbf{B} = (\Phi^{\mathcal{M}})^{\dagger} \mathbf{G}$, with $\Phi^{\dagger}$ the Moore pseudo-inverse of $\Phi$. We stress again that this step is where we shift focus from the spatial to the spectral domain, and corresponds to the dot product blocks in Figure \ref{fig:architecture}.

In the pipeline first introduced in \cite{ovsjanikov2012functional} and then widely used in the follow-up works \cite{ovsjanikov2017computing}, the standard strategy is to compute the functional map $\mathbf{C}$ that optimizes the following energy:

\begin{equation}
\min_{\mathbf{C}} \big\Vert \mathbf{C} \mathbf{A} - \mathbf{B} \big\Vert^2 + \lambda \big\Vert \mathbf{C} \Delta_{\mathcal{M}} - \Delta_{\mathcal{N}} \mathbf{C} \big\Vert^2,
\label{eq:fmap_lap}
\end{equation}
where $\lambda$ is a scalar regularization parameter.

Remark that the optimization problem in Eq. \eqref{eq:fmap_lap} is quadratic in terms of $\mathbf{C}$ and can be solved e.g. via standard  convex optimization techniques. However, in the learning context, we need to differentiate the solution with respect to the spectral features $\mathbf{A}, \mathbf{B}$, which is challenging when $\mathbf{C}$ is computed via an iterative solver. Alternatively, the problem in Eq. \label{eq:fmap_lap} can be written directly in terms of a large least squares system, by vectorizing the matrix $\mathbf{C}$ as was suggested in  \cite{ovsjanikov2012functional}. However, for a $k \times k$ functional map, this leads to a system of size $k^2 \times k^2$ which becomes prohibitive even for moderate values of $k$. To avoid these issues, previous learning-based approaches based on functional maps \cite{litany2017deep,halimi2018self,roufosse2019unsupervised} have only optimized for $\mathbf{C}$ using the first part of the energy in Eq. \eqref{eq:fmap_lap}: $\big\Vert\mathbf{C} \mathbf{A} - \mathbf{B} \big\Vert^2 $. This results in a simple linear system for which the derivatives can be computed in closed form. This has two major limitations, however: first the linear system is only invertible if there are at least $k$ linearly independent feature functions. This condition can easily be violated in practice, especially in the early stages of learning, potentially resulting in a fatal error. Furthermore, the lack of regularization makes the solved-for functional map very sensitive to  inconsistencies in the computed descriptors, which leads to an overall loss of robustness.

In our work we address this problem by using the full energy in Eq. \eqref{eq:fmap_lap} in a fully differentiable way. In particular, we use the fact that the operators $\Delta_{\mathcal{M}}$ and $\Delta_{\mathcal{N}}$ are diagonal when expressed in their own eigen-basis.

Indeed we remark that the gradient of the energy in Eq. \eqref{eq:fmap_lap} vanishes whenever $\mathbf{C}\mathbf{A}\mathbf{A}^T + \lambda \matr{\Delta} \cdot \mathbf{C} = \mathbf{B}\mathbf{A}^T$ , where the operation $\cdot$ represents the element-wise multiplication, and $\matr{\Delta}_{ij} = (\mu^{\mathcal{M}}_j - \mu^{\mathcal{N}}_i)^2$, where $\mu^{\mathcal{M}}_l$ and $\mu^{\mathcal{N}}_l$ respectively correspond to the eigenvalues of $\Delta_{\mathcal{M}}$ and $\Delta_{\mathcal{N}}$. It is then easy to see that this amounts to a separate linear system \emph{for every row} $c_i$ of $\mathbf{C}$ :

\begin{equation}
(\mathbf{A}\mathbf{A}^T + \lambda \text{diag}((\mu^1_j - \mu^2_i)^2)) c_i = \mathbf{A} b_i
\label{lapreg}
\end{equation}

where $b_i$ stands for $i^{th}$ row of $B$.

In total, if $k$ is the number of eigenvectors used for representing the functional map, this operation amounts to inverting $k$ different $k \times k$ matrices. Since inverting a linear system is a differentiable operation, which is already implemented e.g. in TensorFlow, this allows us to estimate the functional map in a robust way, while preserving differentiability. 


\subsection{The supervised spectral loss}
\label{subsec:spec_loss}

Our method also uses a loss with respect to the ground truth functional map in the spectral domain. This is similar to the energy used in \cite{corman2014supervised}, but is different from the loss of the original FMNet work \cite{litany2017deep}, which converted a functional map to a soft correspondence matrix and imposed a loss with respect to the ground truth point-wise map, relying on expensive geodesic distance matrix computation. 


Specifically, calling $\mathbf{C}$ the functional map obtained by the FMap block, and $\mathbf{C}^{gt}$ the ground truth spectral map, our loss is defined as:
$$
l_{\text{spec}} = \big\Vert\mathbf{C} - \mathbf{C}^{gt} \big\Vert^2
$$
As mentioned above, we use a Fröbenius norm to compute the distance between matrices.

It is important to note that whenever a pointwise ground truth map is given it is trivial to convert it to the functional map representation. Conversely, the ground truth spectral map is more general than the point-wise ground truth correspondence. Indeed, with just a few precise landmarks one can recover a functional map accurate enough to make this loss efficient, for instance through the original pipeline of \cite{ovsjanikov2012functional}, but also with more recent follow-up works, such as \cite{ren2018continuous} or \cite{melzi2019zoomout}, which we will further describe as baselines to our method in Section \ref{sec:results}.

This is useful, e.g., in the case of re-meshed datasets. Indeed, complete ground truth correspondences between two shapes of these datasets are not fully known. One can only have access to the (often partial and not bijective) ground truth pointwise map from a template mesh $\mathcal{T}$ to each re-meshed shape $\mathcal{S}_i$. Ecah such map can be converted to a functional map   $\mathbf{C}_i$ and a very good approximation of the spectral ground truth $C^{gt}_{i \rightarrow j}$ between $\mathcal{S}_i$ and $\mathcal{S}_j$ can be expressed as $\mathbf{C}_j^{\dagger} \mathbf{C}_i$.

\subsection{Postprocessing}

Once our model is trained, we can then test it on a pair of shapes and get a functional map between these shapes. This map can either directly be converted to a point to point map, or refined further. We use a very recent and efficient refining algorithm, called ZoomOut \cite{melzi2019zoomout} based on navigating between spatial and spectral domains while progressively inceasing the number of spectral basis functions. This efficient postprocessing technique allows us to get state-of-the-art results, as described in Section \ref{sec:results}.

\subsection{Implementation}

We implemented our method in TensorFlow \cite{tensorflow2015-whitepaper} by adapting the open-source implementation of SURFMNet \cite{roufosse2019unsupervised} and KPConv \cite{thomas2019KPConv}. 

Our feature extraction network is based on a residual convolutional architecture of \cite{thomas2019KPConv}, consisting of 4 convolutional blocks with leaky linear units, with successive poolings and dimension augmentation from 128 to 2048, followed by a 4 up-sampling blocks with shortcuts from corresponding pooling layers, and dimension reduction from 2048 back to 128.
Please see the Supplementary materials, part A, in \cite{thomas2019KPConv} for more details.
Following the pipeline of KPConv, we start with a sub-sampled version of our point clouds with a grid subsampling of step 0.03. 
The pooling layers are therefore obtained with grid samplings of parameters 0.06, 0.12, 0.24 and 0.48.

Similarly to FMNet \cite{litany2017deep} and SURFMNet \cite{roufosse2019unsupervised}, our network is applied in a Siamese way on the two shapes, using the same learned weights for the feature extractor.

In the case of fully automatic spectral methods such as BCICP \cite{ren2018continuous} and ZoomOut \cite{melzi2019zoomout}, or the deep learning based FMNet \cite{litany2017deep, halimi2018self} (supervised or unsupervised) and SURFMnet \cite{roufosse2019unsupervised}, all results are invariant by any rigid transformation of the input shapes. However, in the case of methods using the 3D coordinates of the points to generate information about the input shape, this does not remain true. Consequently, both 3D-CODED \cite{groueix20183d} and our method avoid this dependency through data augmentation to be as close as possible to the generality of fully spectral methods. To that end, assuming the shapes are all aligned on one axis (e.g. on human the natural up axis), both 3D-CODED and our method perform data augmentation by randomly rotating the input shapes along that axis. 


\subsection{Parameters}
In addition to the architecture above, our method has two key hyper-parameters: the size of the functional basis and the regularizer $\lambda$ in Equation \ref{lapreg}. For the size of the basis, we discovered if this number is too high, for instance, with 120 eigenvectors as in FMNet and SURFMNet, it can easily lead to overfitting. However, by reducing this number to 30, the results of SURFMNet on FAUST re-meshed (here reported in Table \ref{table:res1}) go from 0.15 to 4.5. As a consequence, we choose the number of eigenvectors to be 30 in all of our experiments on our method. Regarding the weight $\lambda$ in Equation \eqref{lapreg}, we observed that setting it to $\lambda = 10^{-3}$ helps getting good results while drastically reducing the number of training steps, as pointed out in the ablation study. We use this value throughout all experiments.

\begin{table}
\begin{center}
\begin{tabular}{|l|c|c|c|c|}
\hline
Method $\backslash$ Dataset & F & S & F on S & S on F \\
\hline\hline
BCICP         & 15. & 16. &  *  &  * \\
ZoomOut       & 6.1 & 7.5 &  *  &  * \\
\hline
SURFMNet       & 15. & 12. & 32. & 32.\\
SURFMNet+icp   & 7.4 & 6.1 & 19. & 23.\\
Unsup FMNet    & 10. & 16. & 29. & 22.\\
Unsup FMNet+pmf& 5.7 & 10. & 12. & 9.3\\
\hline
FMNet         & 11. & 17.& 30. & 33.\\
FMNet+pmf     & 5.9 & 6.3 & 11. & 14.\\
3D-CODED      & 2.5 & 31. & 31. & 33.\\
\hline
Ours          & 3.1 & 4.4 & 11. & 6.0\\
Ours+zo       & \textbf{1.9} & \textbf{3.0} & \textbf{9.2} & \textbf{4.3}\\
\hline
\end{tabular}
\end{center}
\caption{Comparative results ($\times 100$) of the different methods on Experiment 1.}
\label{table:res1}
\end{table}

We train our network with a batch size of 4 shape pairs for a number of epochs depending on the number of shapes in the dataset. 
We use a learning rate of .001 and gradually decreasing it to 0.0001 with ADAM optimizer \cite{adam}.

%% file: sections/results.tex
\section{Results}
\label{sec:results}

\subsection*{Datasets}

We test our method on a wide spectrum of human datasets: first, the re-meshed versions of FAUST dataset \cite{bogo2014} containing 100 human shapes in 1-1 correspondence, and of SCAPE \cite{scape}, made publicly available by Ren et al. \cite{ren2018continuous}.
These re-meshed datasets offer significantly more variability in terms of shape structures and connectivity, including for instance point sampling density, making them harder to match for existing algorithms. We also highlight that the SCAPE dataset is slightly more challenging since the shapes are less regular, and two shapes never share the same pose. 
This is not true for FAUST, wherein all the poses present in the test set also exist in the training set, with the variation coming from body type only, making the pose recovery easier at test time.

We also use the re-meshed version of the more recent SHREC'19 dataset \cite{SHREC19}, which, in theory, is the most challenging of the test sets, because of stronger distortions in the poses, the presence of an incomplete shape, and the number of test pairs (430 in total, so two times the number of test pairs of FAUST or SCAPE). 
At last, we also use the generic training dataset of 3D-CODED \cite{groueix20183d}, originally consisting in 230K synthetic shapes generated using Surreal \cite{varol17_surreal}, with the parametric model SMPL introduced in \cite{SMPL_2015}. We use it only for training purposes in our second experiment, to show that our method can generalize well to changes in connectivity, being able to train on a synthetic, very smooth, identical triangulation for the whole training set, and still produce results of excellent quality on re-meshed datasets.

\paragraph{Ablation study}

\comment{
\begin{table}
\begin{center}
\begin{tabular}{|l|c|c|}
\hline
Method & No Ref & Ref \\
\hline\hline
FMNet      & 17. & 13. \\
PointNet   & 18. & 14. \\
Old FMap   & 4.5 & \textbf{1.9} \\
Ours       & \textbf{3.4} & \textbf{1.9} \\
\hline
\end{tabular}
\end{center}
\caption{Comparative results for the different ablations of our method.}
\label{table:resAblationStudy}
\end{table}
}
Our method is built with a number of building blocks, all of which we consider essential to achieve optimal performance. To illustrate this, in the supplementary materials we provide an extensive ablation study of all the key components of our algorithm.

\paragraph{Baselines}

\begin{figure*}[t!]
\begin{center}
\includegraphics[width=0.9\linewidth]{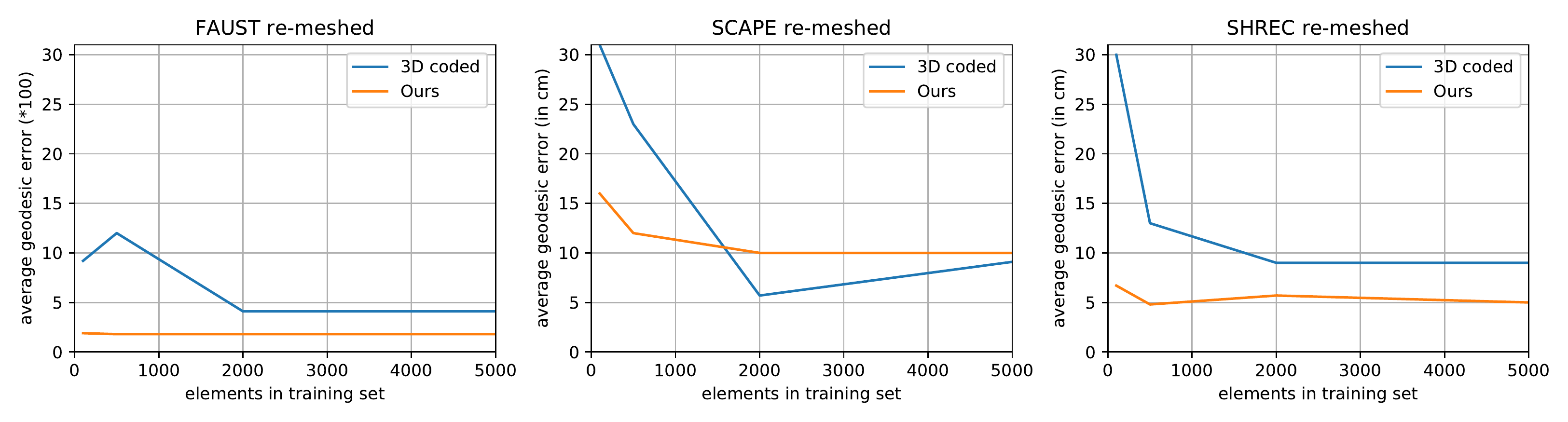}
\end{center}
\vspace{-1mm}
   \caption{Comparison with 3D-CODED while varying training size of SURREAL dataset and simultaneously testing on other datasets.\vspace{-2mm}}
\label{fig:exp2_res}
\end{figure*}

\comment{
\begin{table}
\begin{center}
\begin{tabular}{|l|c|c|c|c|}
\hline
Method $\backslash$ Dataset & F & S & F on S & S on F \\
\hline\hline
BCICP         & 15. & 16. &  *  &  * \\
ZoomOut       & 6.1 & 7.5 &  *  &  * \\
\hline
SURFMNet       & 15. & 12. & 32. & 32.\\
SURFMNet+icp   & 7.4 & 6.1 & 19. & 23.\\
Unsup FMNet    & 10. & 16. & 29. & 22.\\
Unsup FMNet+pmf& 5.7 & 10. & 12. & 9.3\\
\hline
FMNet         & 11. & 17.& 30. & 33.\\
FMNet+pmf     & 5.9 & 6.3 & 11. & 14.\\
3D-CODED      & 2.5 & 31. & 31. & 33.\\
\hline
Ours          & 3.1 & 4.4 & 11. & 6.0\\
Ours+zo       & \textbf{1.9} & \textbf{3.0} & \textbf{9.2} & \textbf{4.3}\\
\hline
\end{tabular}
\end{center}
\caption{Comparative results ($\times 100$) of the different methods on Experiment 1.}
\label{table:res1}
\end{table}
}

We compare our method to several state of the art methods: the first category includes fully automatic methods without any learning component \cite{ren2018continuous, melzi2019zoomout}. These methods are simply evaluated on the test sets without any training. The second category includes FMNet \cite{litany2017deep} and its unsupervised versions, referred to as Unsup FMNet \cite{halimi2018self} and SURFMNet \cite{roufosse2019unsupervised}, with and without post-processing (PMF \cite{vestner2017product} for FMNet, and standard functional map refinement \cite{ovsjanikov2012functional}, referred to as ICP, for SURFMNet). All these variants of FMNet give similar results, but SURFMNet is the only one to train within a few hours, without requiring too much space. This is due to the fact SURFMNet only operates in the spectral domain, in contrast to other methods. Lastly, we compare to the supervised 3D-CODED \cite{groueix20183d}, described earlier in more details in Section \ref{subsec:fmaps}.
For conciseness, we refer to our method as Ours in the following text. We show our results with and without ZoomOut \cite{melzi2019zoomout} refinement, referred to as ZO, in order to prove that our method stands out even without post processing. We compare these different methods in two main settings named Experiment 1 and Experiment 2 below.

\paragraph{Experiment 1} consists of evaluating the different methods in the following setting: we split FAUST re-meshed and SCAPE re-meshed into training and test sets containing 80 and 20 shapes for FAUST, and 51 and 20 shapes for SCAPE. We obtain results for training and testing on the same dataset, but also by testing on a different dataset. For instance, by training on SCAPE re-meshed train set and testing on FAUST re-meshed test set. This experiment aims at testing the generalization power of all methods to small re-meshed datasets, as well as its ability to adapt to a different dataset at test time.

\paragraph{Experiment 2} consists of sampling 100, 500, 2000, and 5000 shapes from the SURREAL dataset to be used for training. We then test the trained models on the test sets of FAUST re-meshed, SCAPE re-meshed, and SHREC19 re-meshed. This experiment aims at testing the robustness and generalization power of the different methods in the presence of varying amounts of training data, as well as adaptability to train on a perfect synthetic triangulations and still get results on challenging re-meshed shapes.

\subsection*{Quantitative results}
\label{subsec:quant_res}

To evaluate the results, we use the protocol introduced in \cite{kim11}, where the per-point-average geodesic distance between the ground truth map and the computed map is reported. All results are multiplied by 100 for the sake of readability. 

As we can see in Table \ref{table:res1}, our method performs the  best overall on Experiment 1. Fully automatic methods do not provide competitive results compared to the learning methods (except on crossed settings because they did not train on anything and are thus not influenced by the training shapes). As reported in the Section \ref{sec:background}, this highlights that hand-crafted features can easily fail.
It is noticeable that spectral methods (FMNet variations, and Ours as a hybrid method) get reasonable, or even good results in our case, with these small datasets. In comparison, 3D-CODED seems to fail in almost all cases. It is remarkable that it can learn on such a small dataset as the training set of FAUST re-meshed. One explanation for that is that FAUST contains the same set of poses in the test set as in the train set.

Contrary to other baselines, our method gives good results on all settings, \textit{even without refinement}, showing good resilience to a really low number of shapes, even with re-meshed geometry. We would like to stress that no other method is able to achieve such a generalization with this low number of shapes.

For a fair comparison with 3D-CODED, we complete our study with a second experiment, in which the training set is now made of the same shapes 3D-CODED uses for training in their paper, namely SURREAL dataset. The aim of this experiment is to further showcase the generalization power of our method when compared to 3D-CODED. First, by training on a very smooth synthetic dataset, on which previous fully spectral methods tend to easily overfit due to the obvious mismatch in triangulation in training and test set. Our second goal is to observe the dependence of different methods on size of the training set.

\begin{figure*}[t!]
\begin{center}
\includegraphics[width=0.9\linewidth]{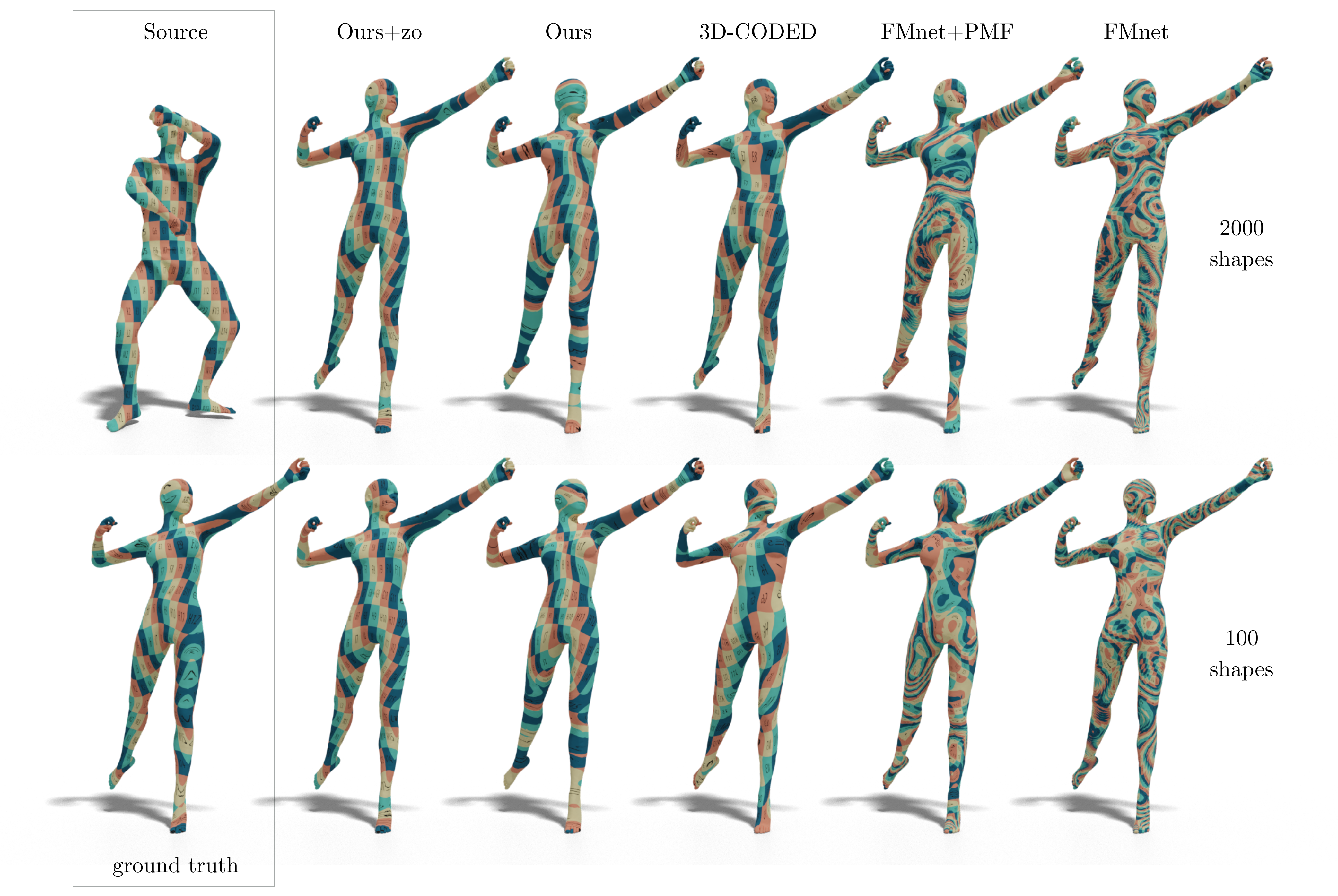}
\end{center}
   \caption{Qualitative results obtained with texture transfers for the different methods on Experiment 2, training on two different numbers of shapes in the SURREAL dataset, and testing on SHREC re-meshed shapes.\vspace{-2mm}}
\label{fig:exp2_res_quali}
\end{figure*}

We report the results (multiplied by 100) of 3D-CODED and Our method in Figure \ref{fig:exp2_res}, as they are the only two competitive algorithms in Experiment 2. These results once again demonstrate that our method can achieve impressive results even with a low number of training shapes. On SHREC re-meshed, we achieve state of the art results with an average error of \textbf{0.048} with only 500 training shapes. We provide additional quantitative comparisons in the supplementary materials.

It can be observed in Figure \ref{fig:exp2_res} that our results are consistent and unaltered even with the drop in number of training shapes. 
3D-CODED, on the other hand, always suffers from a reduced training set. 


\vspace{-2mm}
\subsection*{Qualitative results}
\label{subsec:qualit_res}

In Figure \ref{fig:exp2_res_quali} we show the results of our method (with and without ZoomOut refinement \cite{melzi2019zoomout}), 3D-CODED \cite{groueix20183d}, FMNet \cite{litany2017deep} (with and without PMF refinement \cite{vestnerefficient}), trained on respectively 2000 and 100 shapes, as presented in Experiment 2, via texture transfer.

With 2000 training shapes, both our method and 3D-CODED lead to good or even excellent texture transfers, while fully spectral methods fail due to the change of connectivity from training to test set. 
However, with only 100 training shapes, 3D-CODED fails to get a good reconstruction in many cases, leading to bad texture transfer as in Figure \ref{fig:exp2_res_quali}.
This highlights the fact that our method performs better than any other existing method when only a few training shapes are provided.


%% file: sections/conclusion.tex
\section{Conclusion, Limitations \& Future Work}
\label{sec:conclusion}

We presented a method for improving the robustness and reducing overfitting in learning shape correspondences. Key to our approach is a hybrid network structure, made of a raw-data feature extractor that learns descriptors on a pair of shapes, and a novel robust functional map layer. Our network can thus operate in both the spectral and the spatial domain, thus taking advantages of both representations.

Our approach has several limitations: first, as a supervised method it requires at least partial correspondences (as discussed in Section \ref{subsec:spec_loss}) between the training shapes. Also, it requires data augmentation to be able to predict non-aligned shapes, which can be costly and unstable.

In the future, we plan to work towards an unsupervised spectral loss, similar in spirit to SURFMNet \cite{roufosse2019unsupervised}, while avoiding the symmetry ambiguity problem. We also plan to try other, invariant feature extractors such as \cite{Guerrero2018PCP}, or \cite{poulenard2019SPHNet} to avoid data augmentation.

\paragraph{Acknowledgements}
This work was supported by KAUST OSR Award No. CRG-2017-3426, a gift from Nvidia and the ERC Starting Grant No. 758800 (EXPROTEA). We are grateful to Jing Ren, Simone Melzi and Riccardo Marin for the help with re-meshed shapes, and SHREC'19 dataset. 

%% file: sections/supplementary_material.tex
\subsection{Additional details on KPConv \cite{thomas2019KPConv}}
Here, we review briefly KPConv method and describe the architecture we used in our implementation.

The input to this network is a 3D point cloud equipped with a signal, such as the 3D coordinates of the points. Let $\mathcal{P} \in \mathbb{R}^{N \times 3}$ be a point cloud in $\mathbb{R}^3$. Let $\mathcal{F} \in \mathbb{R}^{N \times D}$ be a $D$-dimensional feature signal over $\mathcal{P}$.

The goal of point cloud convolutional networks is to reproduce the architecture of convolutional neural networks on images. It boils down to transferring two key operations on the point cloud structure: the convolution and the pooling operators.

First, we define a convolution between $\mathcal{F}$ and a kernel $g$ at point $x \in \mathbb{R}^3$. Since we only want a signal over the point cloud at each layer, we only need these convolutions at $x \in \mathcal{P}$.

The kernel will be defined as a local function centered on $0$ depending on some learnable parameters, taking a $D$-dimensional feature vector as input and yielding a $D'$-dimensional feature vector.

More specifically, the kernel is defined in the following way : let $r$ be its radius of action. Let $\mathcal{B}_3^r$ be the corresponding 3d ball. Let $K$ be the number of points, thus the number of parameter matrices in this kernel. Let $\{z_k | k < K \} \subset \mathcal{B}_3^r$ be these points, and $\{W_k | k < K \} \subset \mathcal{M}_{(D, D')}(\mathbb{R})$ be these matrices. Then the kernel $g$ is defined through the formula :

$$
g(y) = \sum_{k<K} h(y,z_k)W_k
$$

where we simply set $h(y,z) = \max (0, 1-\frac{\| y- z \|}{\sigma}) $, so each point of the kernel has a linear influence of range $\sigma$ around it.

Then the convolution simply becomes :

$$
(\mathcal{F} * g) (x) = \sum_{i | x_i \in \mathcal{B}_3^r} g(x_i - x) f_i 
$$

Where the learnable parameters are the matrices $W_k$. We set the points $z_k$ of the kernel to be uniformly organized in $\mathcal{B}_3^r$, so as to better encompass the variations of the convoluted signal at a given point of the point cloud, and a given scale (see \cite{thomas2019KPConv} supplementaries, Section B for more details).


For the pooling operator, we use a grid sampling that allows us to get the point cloud at an adjustable density. The network can then build hierarchical features over the point clouds by both adjusting the radius of influence of its kernels and the density of the mesh they are performed upon.

Once these two operations are set up, it is easy to build a convolutional feature extractor over the point cloud $\mathcal{P}$. In our work, we use the following architecture :
\begin{itemize}
    \item Four strided convolutional blocks, each down-sampling the point cloud to half its density, and taking the feature space to another feature space (corresponding to higher-level features) two times larger.
    \item Four up-sampling layers, taking the signal back on the whole point cloud, through skipping connections followed by 1D convolutions. 
\end{itemize}


\subsection{Ablation study}

\begin{table}
\begin{center}
\begin{tabular}{|l|c|c|}
\hline
Method & No Ref & Ref \\
\hline\hline
FMNet      & 17. & 13. \\
PointNet   & 18. & 14. \\
Old FMap   & 4.5 & \textbf{1.9} \\
Ours       & \textbf{3.4} & \textbf{1.9} \\
\hline
\end{tabular}
\end{center}
\caption{Comparative results for the different ablations of our method.}
\label{table:resAblationStudy}
\end{table}

This section presents the extensive ablation study of all the vital components of our algorithm.

\begin{figure*}[t!]
\begin{center}
\includegraphics[width=0.45\linewidth]{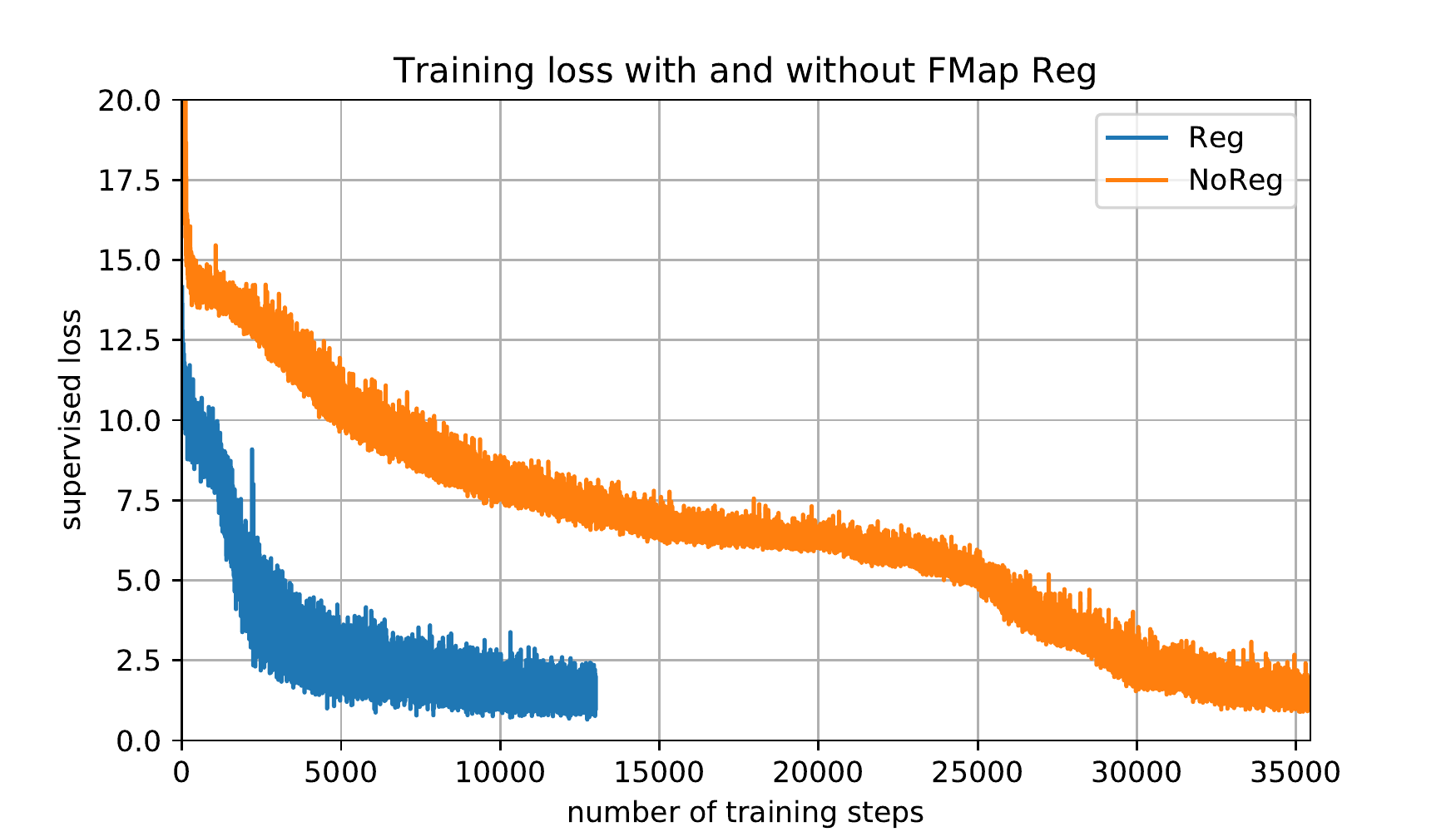}
\includegraphics[width=0.45\linewidth]{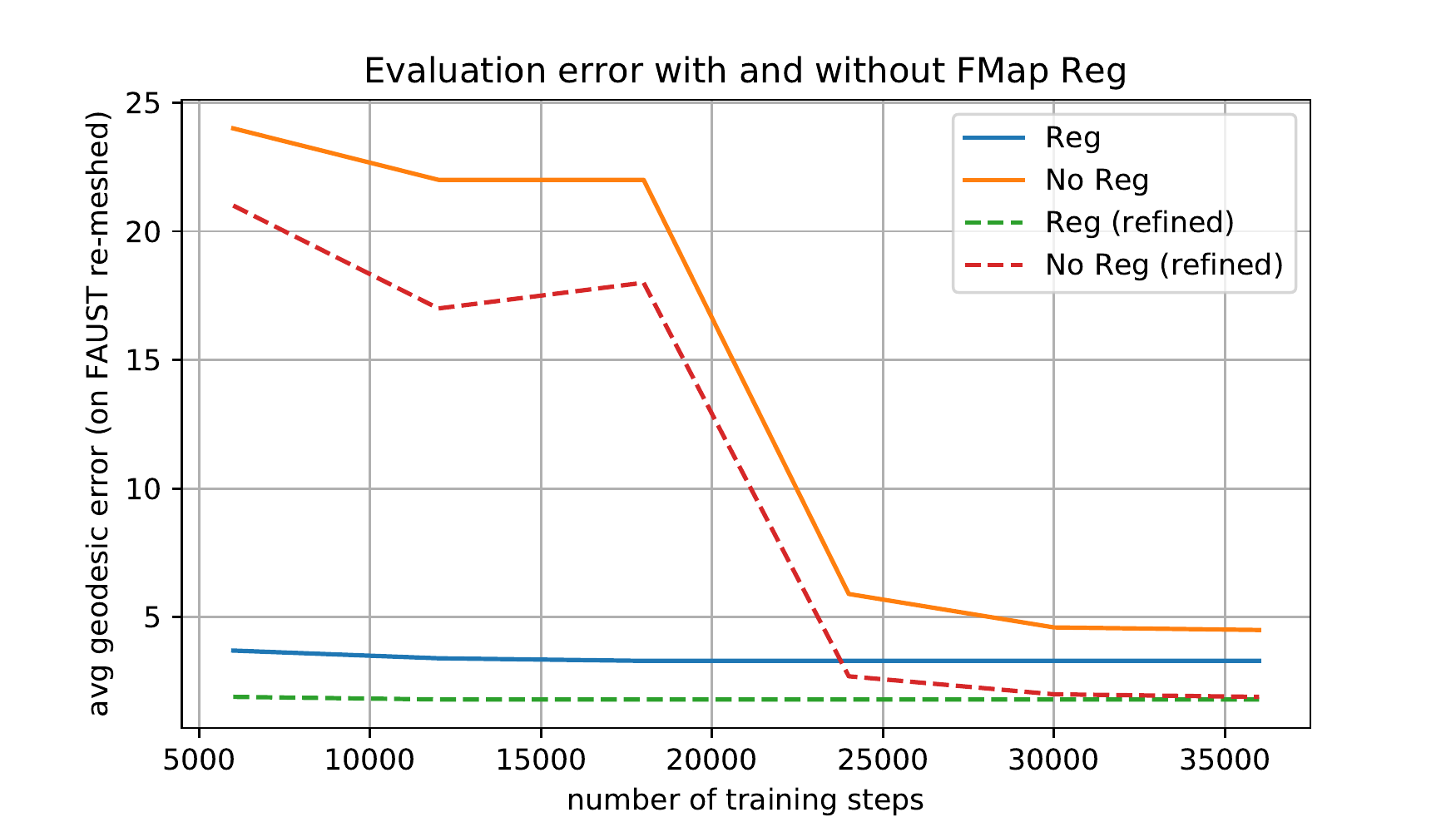}
\end{center}
   \caption{Comparison of convergence speed with and without Laplacian Regularization in the FMap block. \textit{Left}: Training loss evolution, \textit{Right}: Evolution of geodesic error on test set with the number of epochs. Notice how the regularized fmap layer helps drastically with the convergence speed. It gives optimal results within only 500 epochs.}
\label{fig:ab_study_reg}
\end{figure*}

We train all these ablations on 100 random shapes among the 230K proposed by 3D-CODED, as in experiment 2. We test them on the 20 test shapes of FAUST re-meshed, so that the connectivity differs from train to test. The different parts to ablate are :
\begin{itemize}
    \item The point cloud feature extractor : as explained in the previous sections, it learns descriptors from raw data without relying too much on connectivity. The first ablation consists of our method, but with FMNet \cite{litany2017deep} feature extractor instead of ours. Similar to FMNet, we use SHOT \cite{shot} descriptors. However, we use the same number of eigenvectors as in our method, namely 30. As a general remark, we noticed that lowering this number can often help prevent overfitting in the case of FMNet-based architectures.
    \item The choice of KPConv \cite{thomas2019KPConv}. The second ablation study replaces KPConv sampling and feature extractor block with that of PointNet \cite{qi2017pointnet}. For this ablation, we use random sampling to 1500 points instead of KPConv grid sampling. Indeed, grid sampling does not provide any guarantee on the number of points after sampling, so it can only be used in a network built to overcome this issue, with batches of adaptable size, which is not the case of PointNet. 
    \item The regularized functional map layer. This third ablation simply consists in replacing our regularized functional map layer by the old functional map layer originally introduced by FMNet. Our layer is in theory mildly heavier than the original one, but in practice for less than 50 eigenvectors the computation times remain the same.
    \item Lastly, we remove the post-processing refinement step. Here we show the results of every ablation with and without refinement, thus proving it helps in getting better results. As a refinement method, we use the state-of-the-art ZoomOut \cite{melzi2019zoomout}, as mentioned in the main manuscript.
\end{itemize}

Table \ref{table:resAblationStudy} shows the ablation study of our method. It demonstrates the importance of all individual blocks and ascertains that all these components are needed to achieve optimal performance with our solution.

However, just looking at the results of the ablation study one does not see the importance of the FMap Reg addition. To prove its efficiency, we compare the learning and evaluation curves of our method, with and without this addition. As can be seen in Figure \ref{fig:ab_study_reg}, the models converge much faster with our regularized functional map layer. The models are trained on 100 shapes of the surreal dataset of 3D-CODED as in Experiment 2, and tested on FAUST re-meshed.

In addition, our regularized functional map layer is more robust, and does not result in a Cholesky Decomposition fatal error when computing the spectral map. In comparison, the previous functional map layer gave that fatal error in some experiments, and the model had to be relaunched.

Graphically, as reported in the original functional map paper \cite{ovsjanikov2012functional}, a natural functional map should be funnel shaped. Our regularized functional map layer naturally computes maps that almost commute with the Laplacians on the shapes. These maps will naturally be close to diagonal matrices (as are funnel shaped maps) in the eigenbasis of the Laplacians, thus reducing the space of matrices attainable by this layer. We believe it helps the feature extractor block focus on setting the \textit{diagonal} coefficients of the functional map as in the ground truth, rather than on trying to get its funnel shape in the first thousand iterations, which is what a model with the original functional map layer does.

\subsection{More quantitative results}

\begin{figure*}[t!]
\begin{center}
\includegraphics[width=0.31\linewidth]{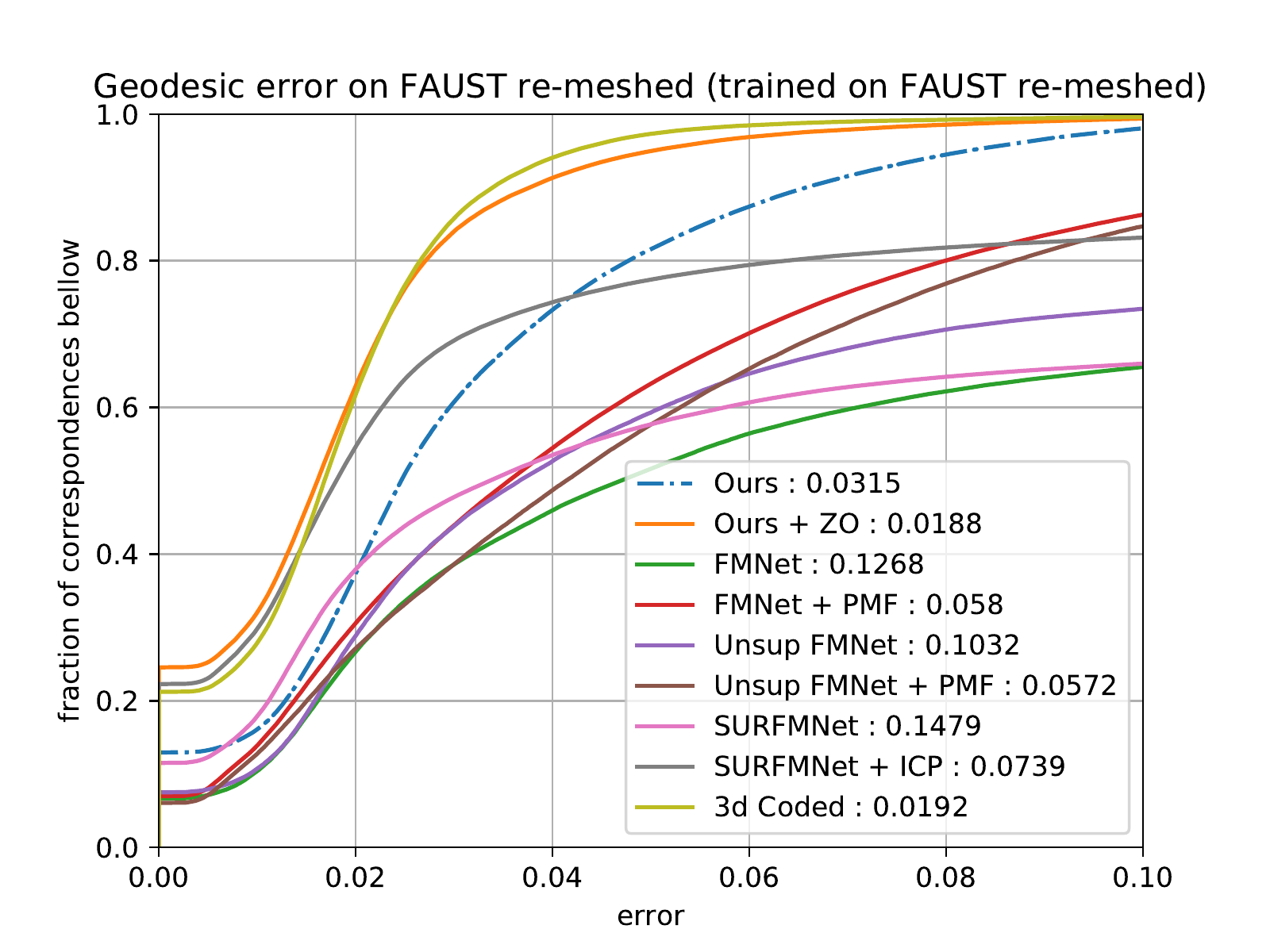}
\includegraphics[width=0.31\linewidth]{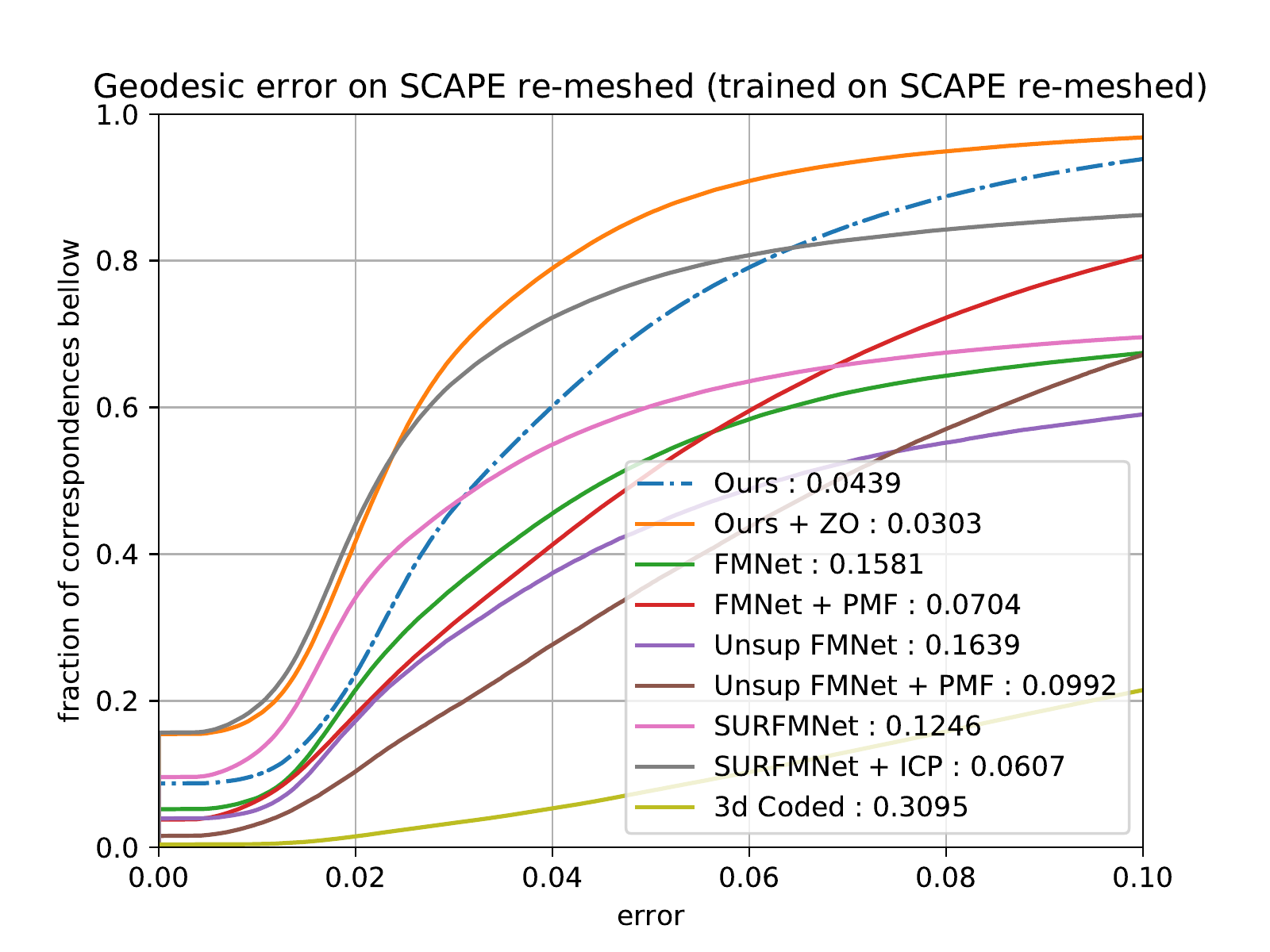}
\end{center}
\begin{center}
\includegraphics[width=0.31\linewidth]{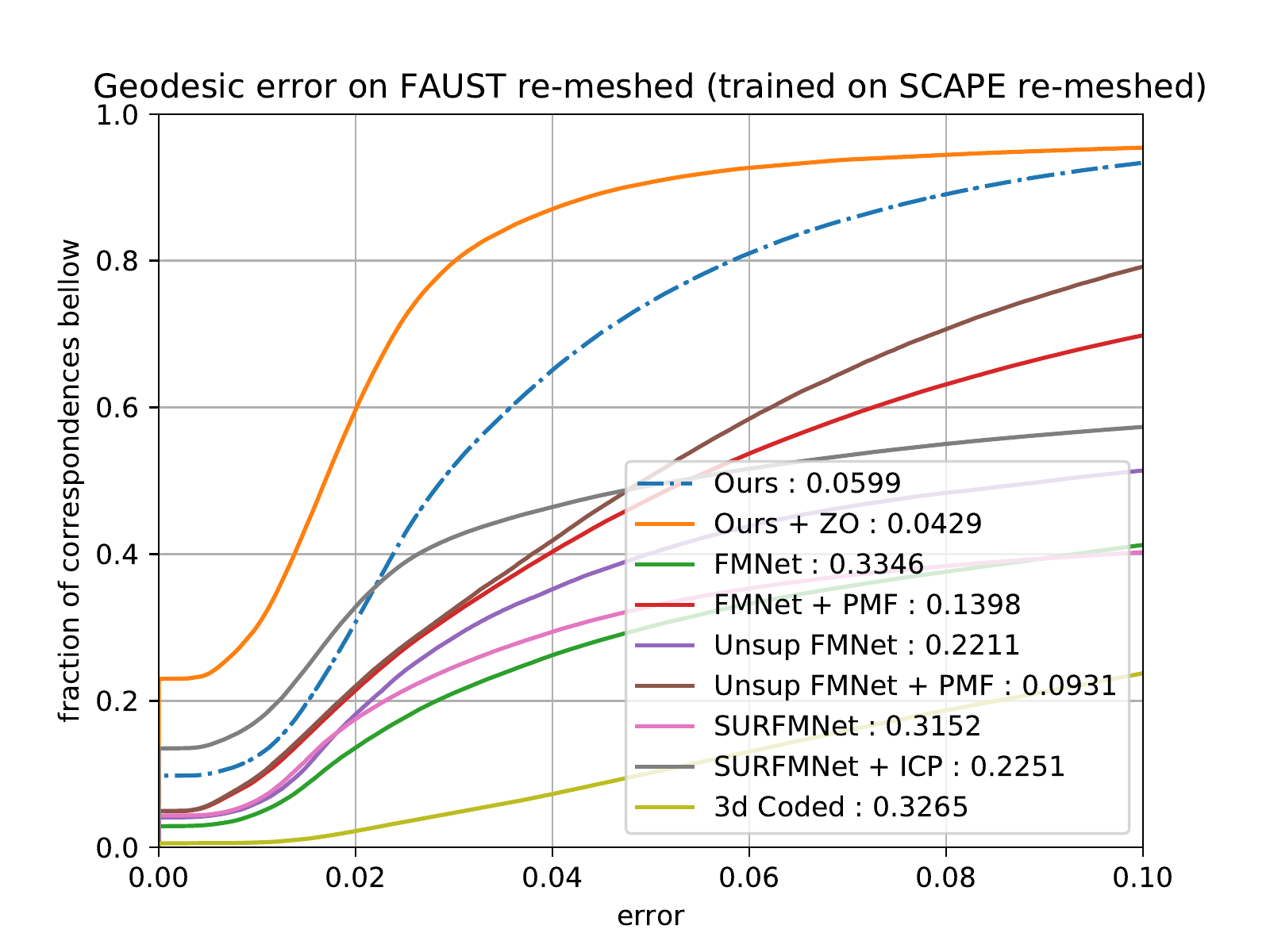}
\includegraphics[width=0.31\linewidth]{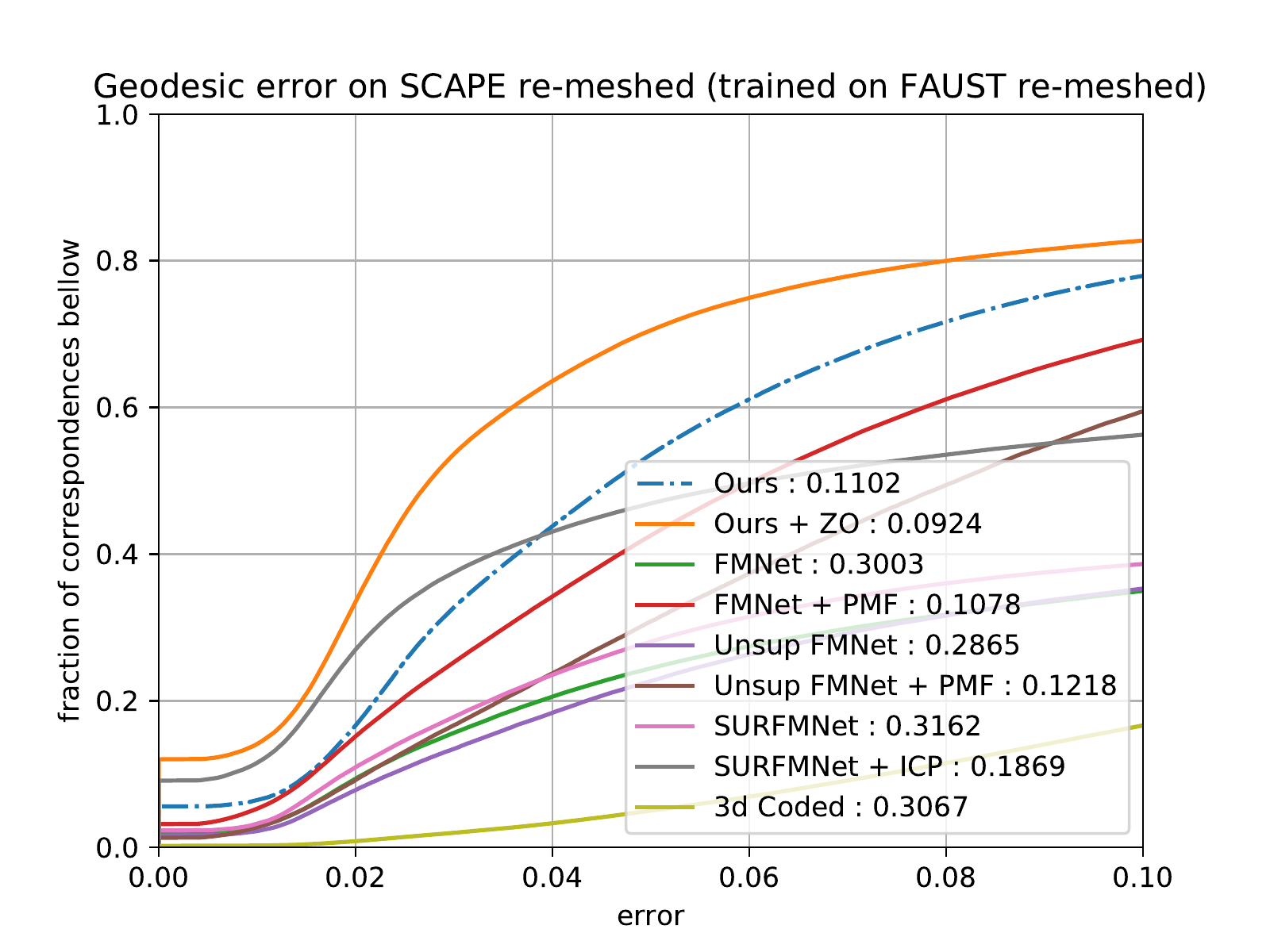}
\end{center}
   \caption{Quantitative results of the different methods using the protocol introduced in \cite{kim11}, on all the settings of Experiment 1}
\label{fig:pr_graphs_1}
\end{figure*}

\begin{figure*}[h]
\begin{center}
\includegraphics[width=0.31\linewidth]{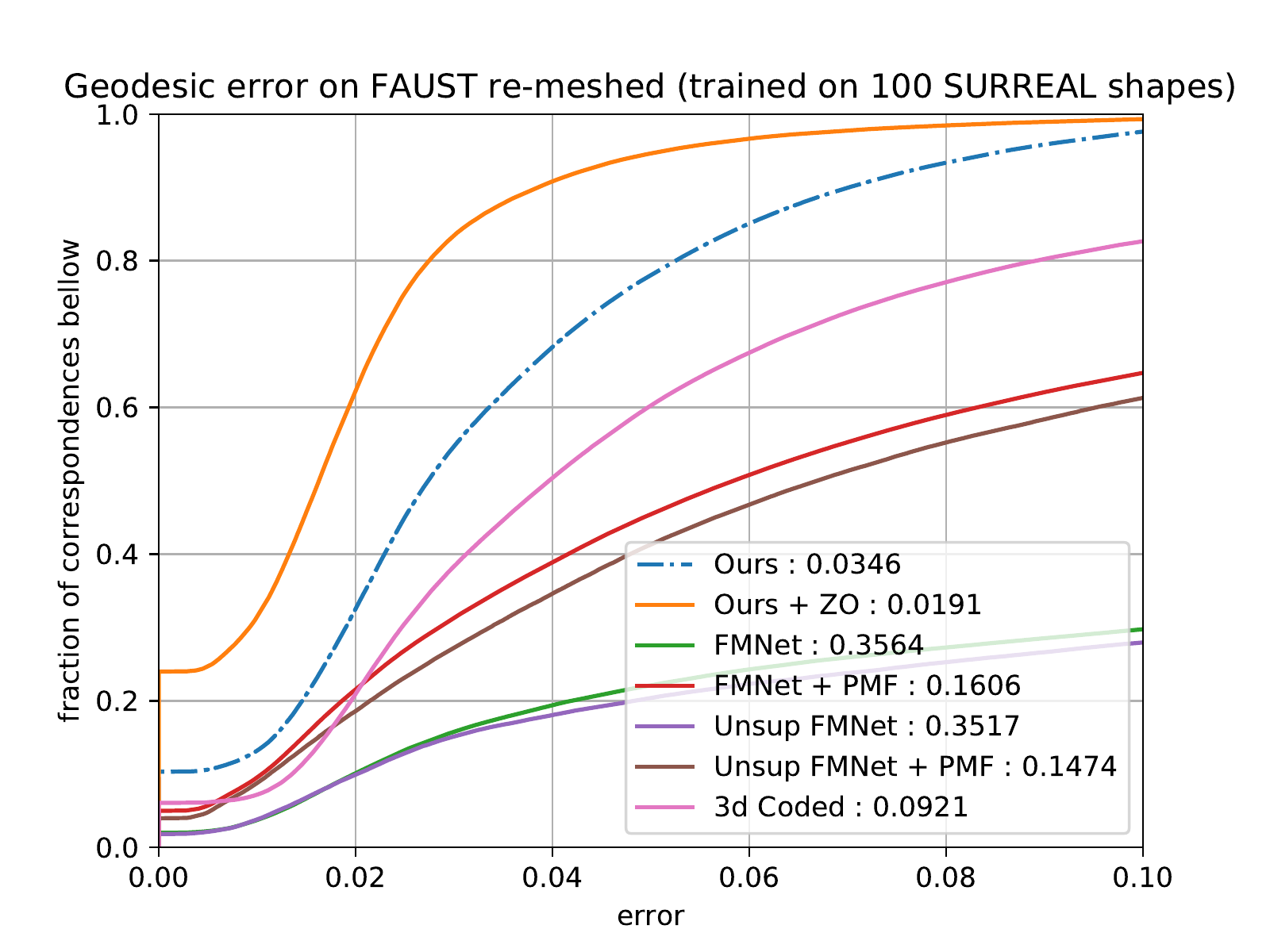}
\includegraphics[width=0.31\linewidth]{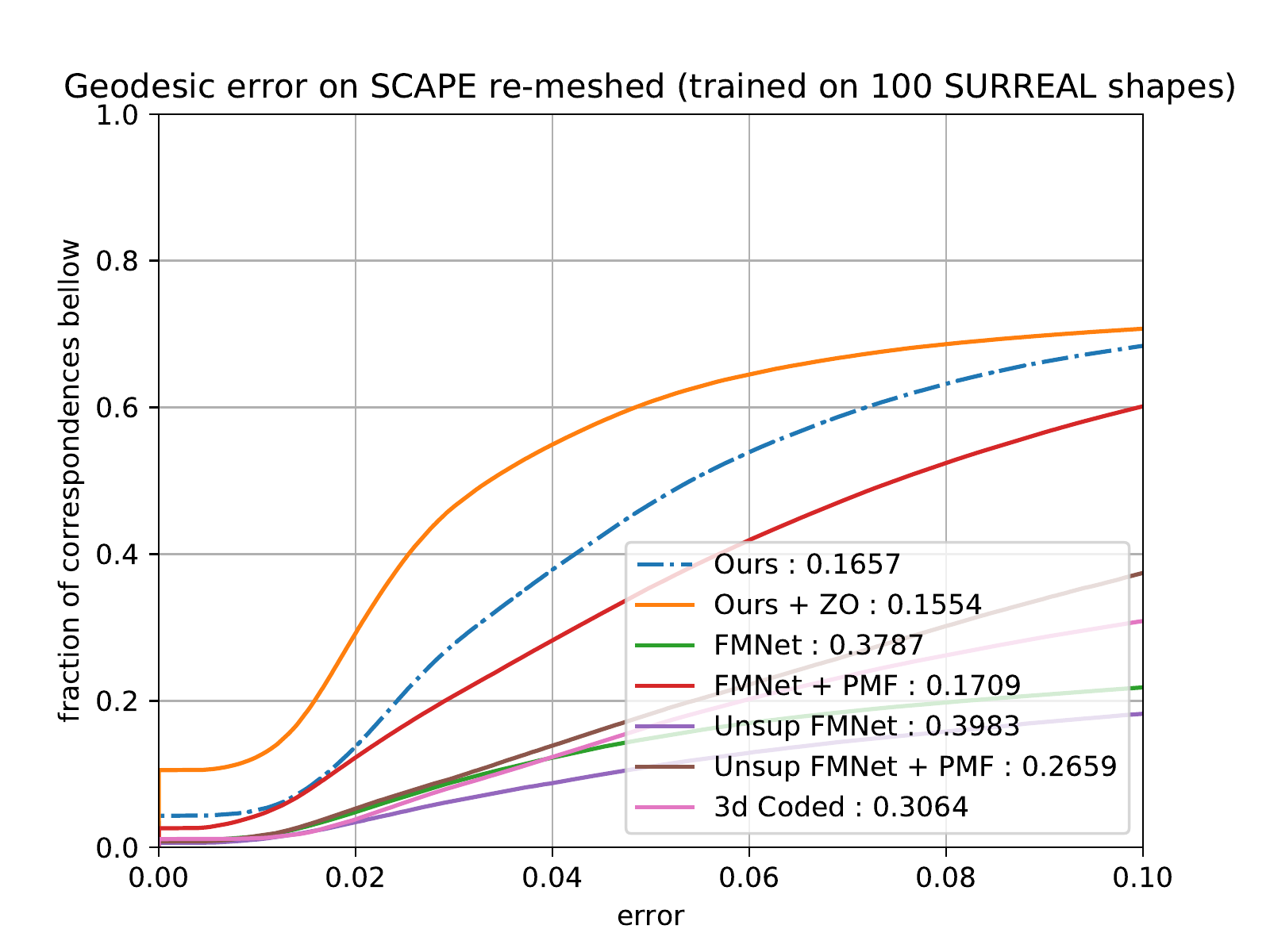}
\includegraphics[width=0.31\linewidth]{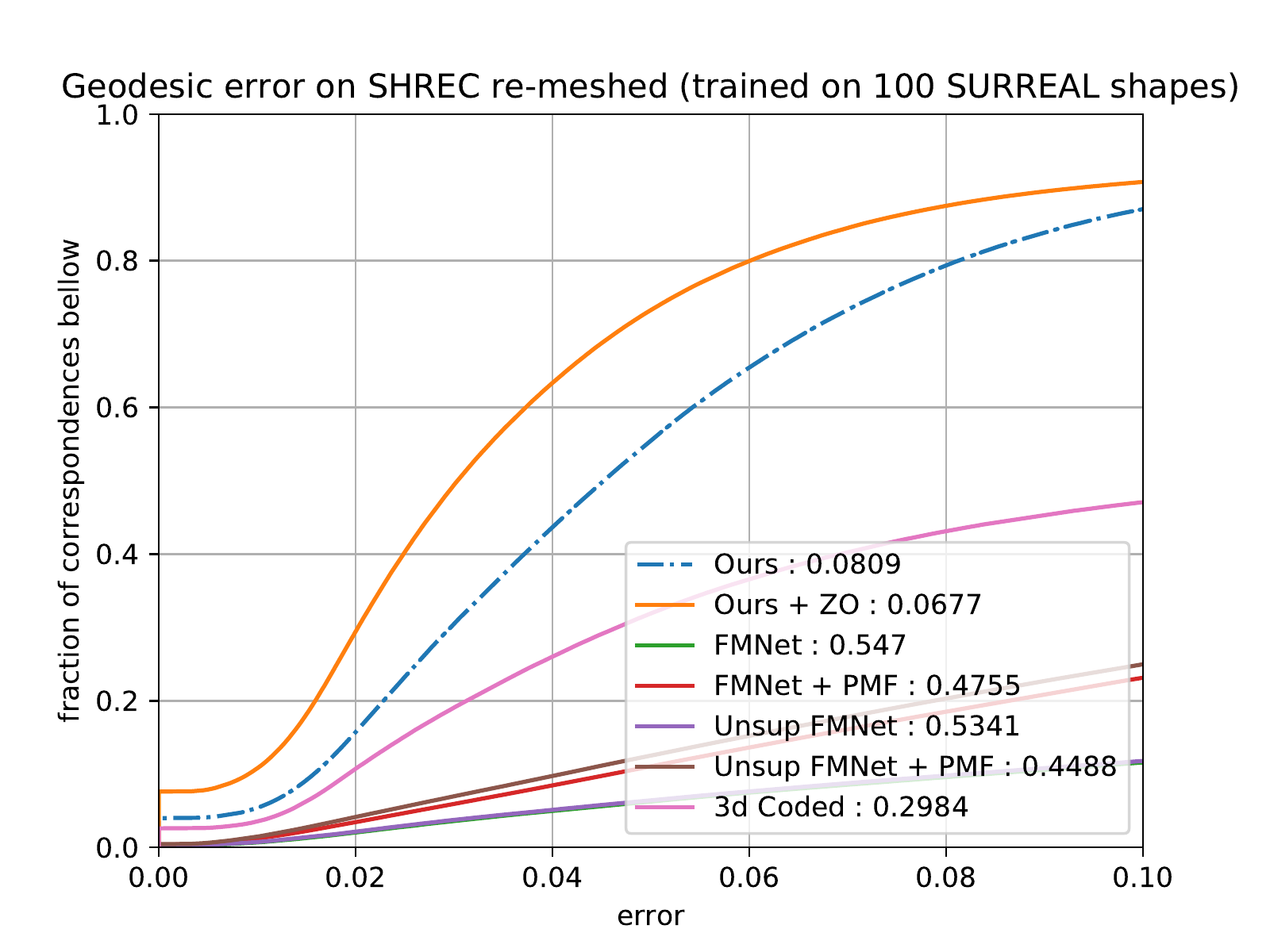}
\includegraphics[width=0.31\linewidth]{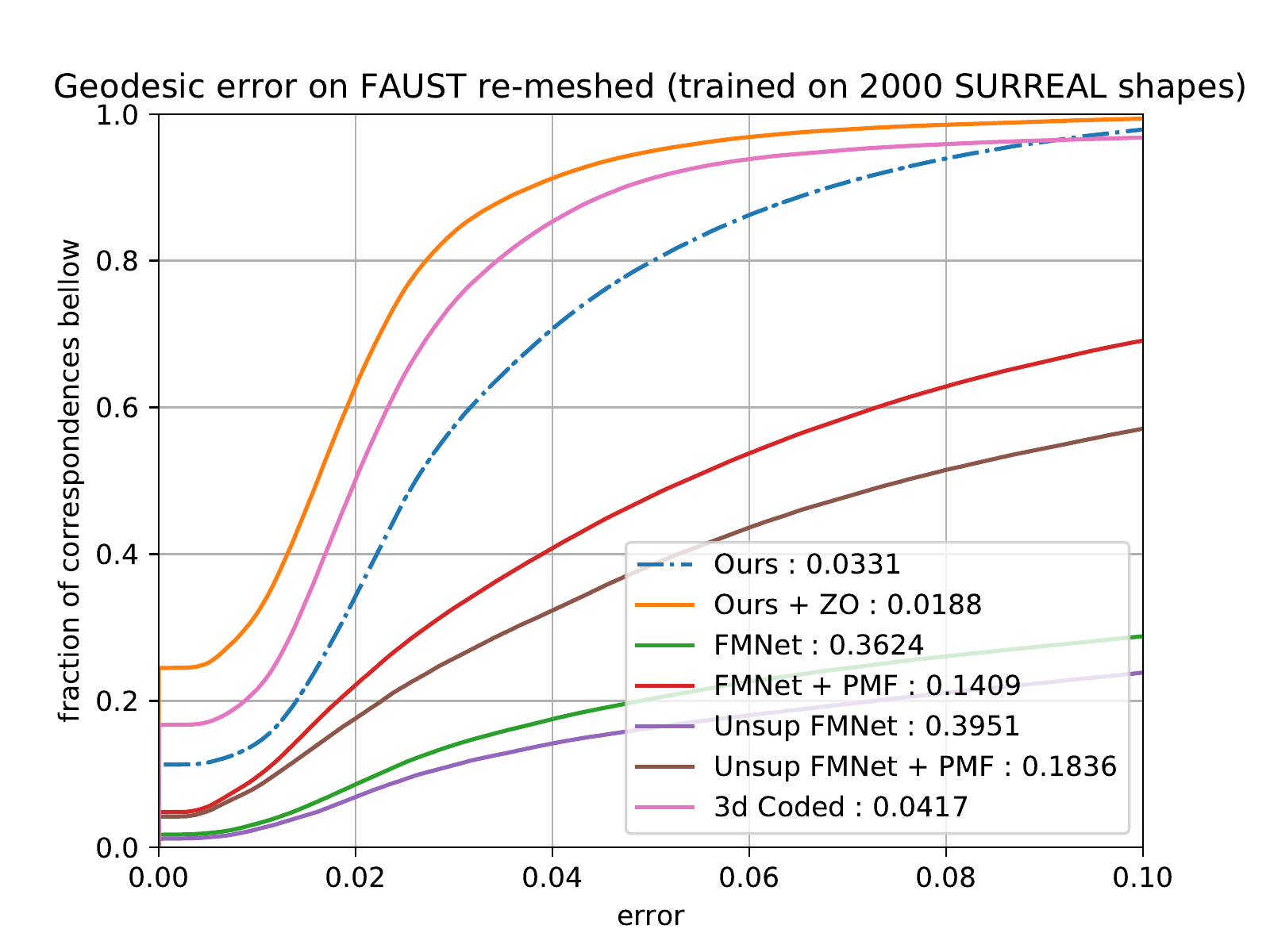}
\includegraphics[width=0.31\linewidth]{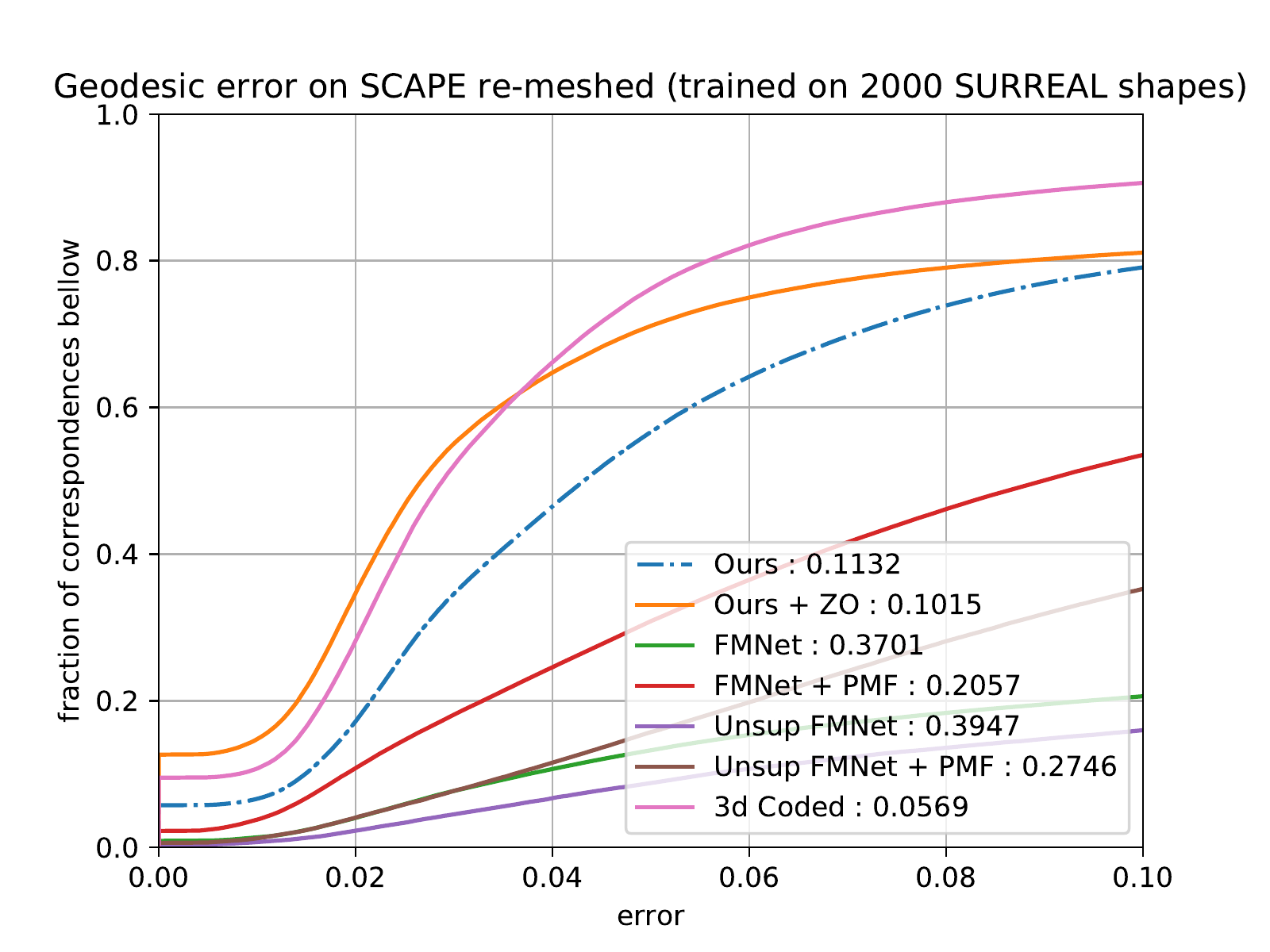}
\includegraphics[width=0.31\linewidth]{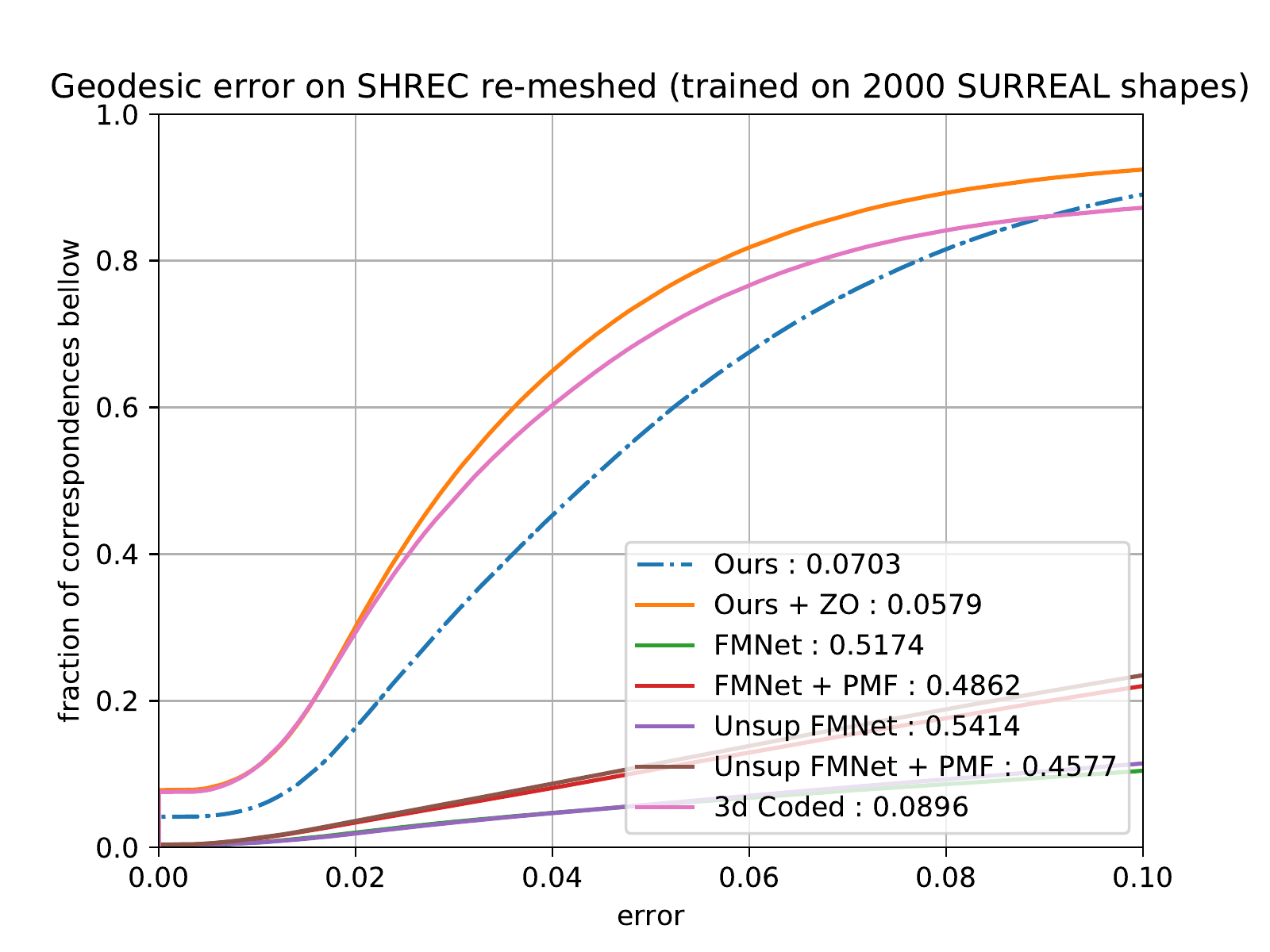}
\end{center}
   \caption{Quantitative results of the different methods using the protocol introduced in \cite{kim11}, on two of the settings of Experiment 2. Top row: 100 shapes (low number). Bottom row: 2000 shapes (high number).}
\label{fig:pr_graphs_2}
\end{figure*}

Figures \ref{fig:pr_graphs_1} and \ref{fig:pr_graphs_2} summarize the accuracy obtained by our method and some baselines on the different settings of the two experiments we conducted, using the evaluation protocol introduced in \cite{kim11}. Note that in all cases but one (trained on 2000 shapes of SURREAL, tested on SCAPE re-meshed), our network achieves the best results even compared to the state-of-the-art methods. As explained more thoroughly in the main manuscript, this proves our method is able to learn point cloud characterizations with only a small amount of data, and by projecting these descriptors in a spectral basis can retrieve accurate correspondences from them. Our method does not need any template and is thus more general than 3D-CODED, in addition to the fact that it trains faster and does not need a big training set.

\begin{figure*}[t]
\begin{center}
\includegraphics[width=1\linewidth]{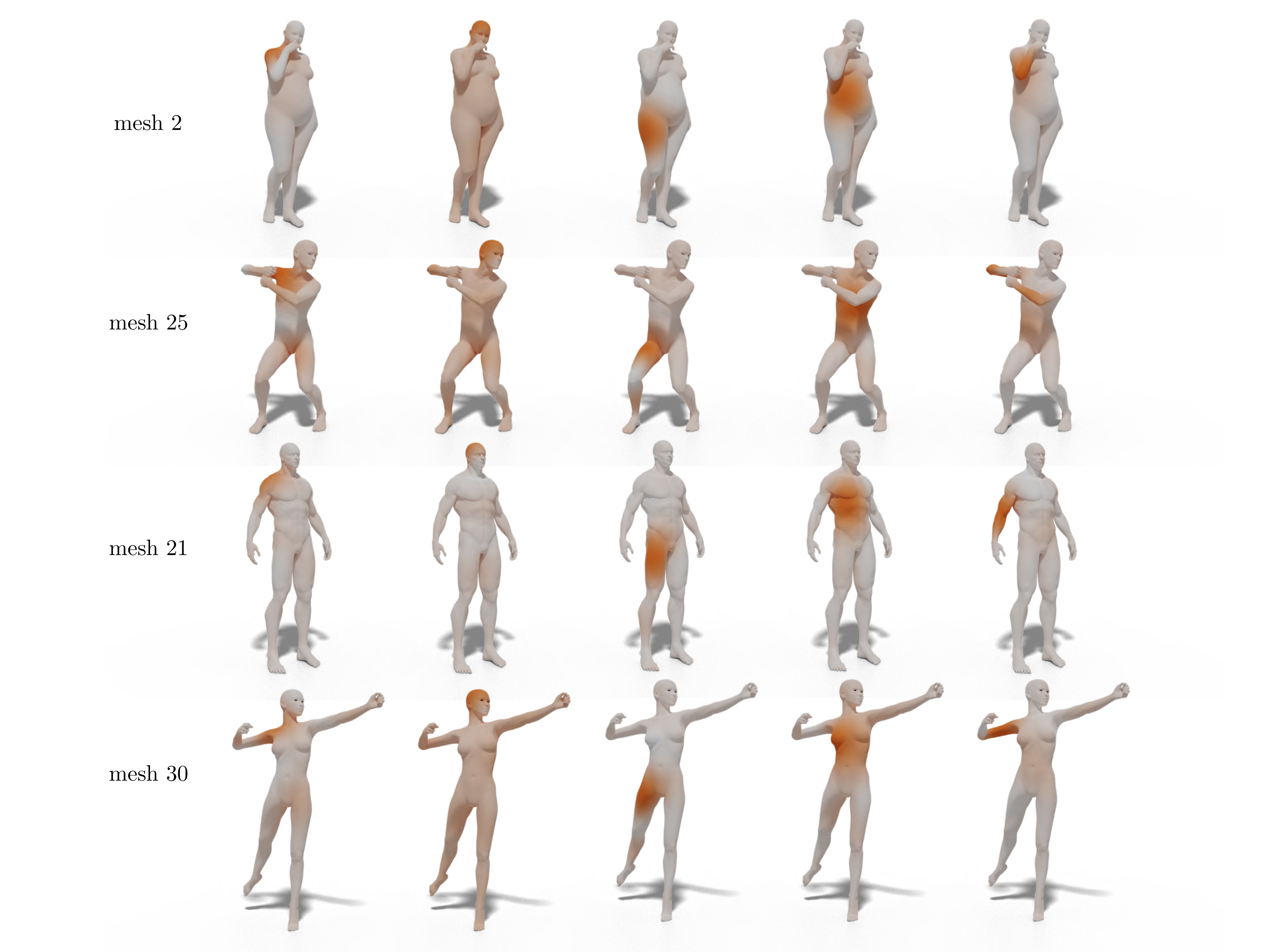}
\end{center}
   \caption{Visualization of spectral descriptors learned by our method (with 2000 surreal shapes) on a test pair of SCAPE re-meshed. The source shape is shown in the first row, and the target shape in the bottom row. Notice how the descriptors are localized and seem to highlight one specific part of the body (first column for shoulder, second for scalp, third for right thigh, fourth for right side of the torso, fifth for elbow).\vspace{-2mm}}
\label{fig:descs}
\end{figure*}

We even believe the superiority of our method with a low number of training shapes shapes is partially due to this fact that 3D-CODED uses a template and operates in the spatial domain, unlike our approach which is template-free, and partly operates in the spectral domain, making it easier to adapt to any new category of 3D shapes.

The relatively low performance of our method on SCAPE in Experiment 2 (see Figure \ref{fig:pr_graphs_2}) is due to the presence of back-bent shapes in this dataset. These shapes are seen by the network through their truncated spectral approximation, as discussed in section \ref{sec:Viz_desc}, making it unable to exploit refined features such as the face or hands, that could help getting descriptors able to differentiate left from right. Consequently, as there are no back-bent shapes in the training sets of this experiment, these shape are often mapped with a left-to-right symmetry, resulting in a huge error for these particular shapes, increasing the mean error for the whole SCAPE test set.

\subsection{Visualization of some descriptors learned by our method}
\label{sec:Viz_desc}

\begin{figure*}[t]
\begin{center}
\includegraphics[width=0.9\linewidth]{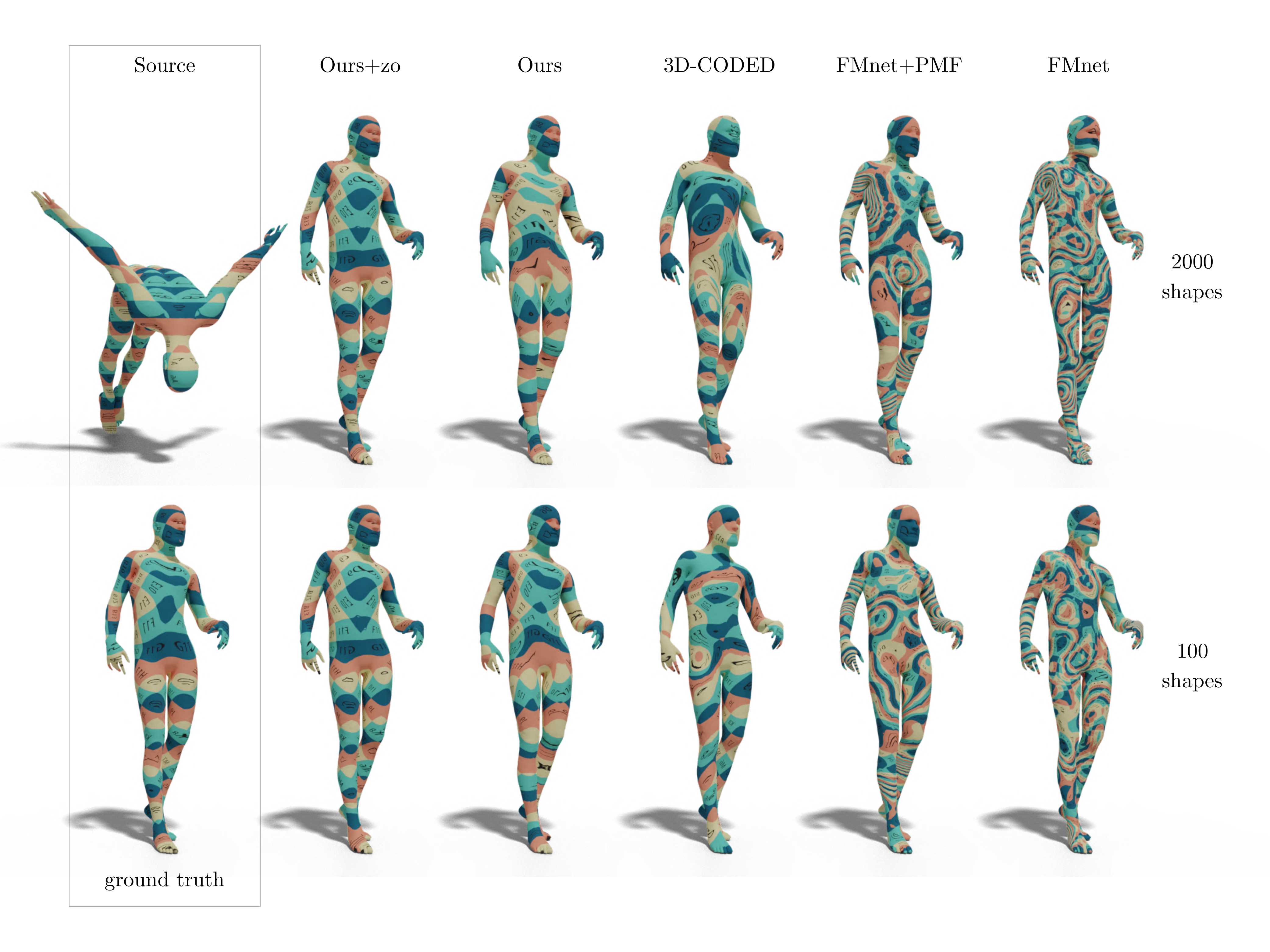}
\end{center}
   \caption{Qualitative results on Experiment 2 through texture transfer, showing cases where our method is the only one that can give good correspondence with only 100 training shapes.}
\label{fig:shrec_maps}
\end{figure*}

Our method aims at building descriptors on both input shapes (that are often labeled source and target shapes) from their raw point cloud data. These descriptors are then projected on the eigen basis of the respective Laplace-Beltrami operators of the source and the target shapes. We output these projections, that we call spectral descriptors, and we visualize some of them in Figure \ref{fig:descs}.

\begin{figure*}[t]
\begin{center}
\hspace{5cm} \includegraphics[width=1\linewidth]{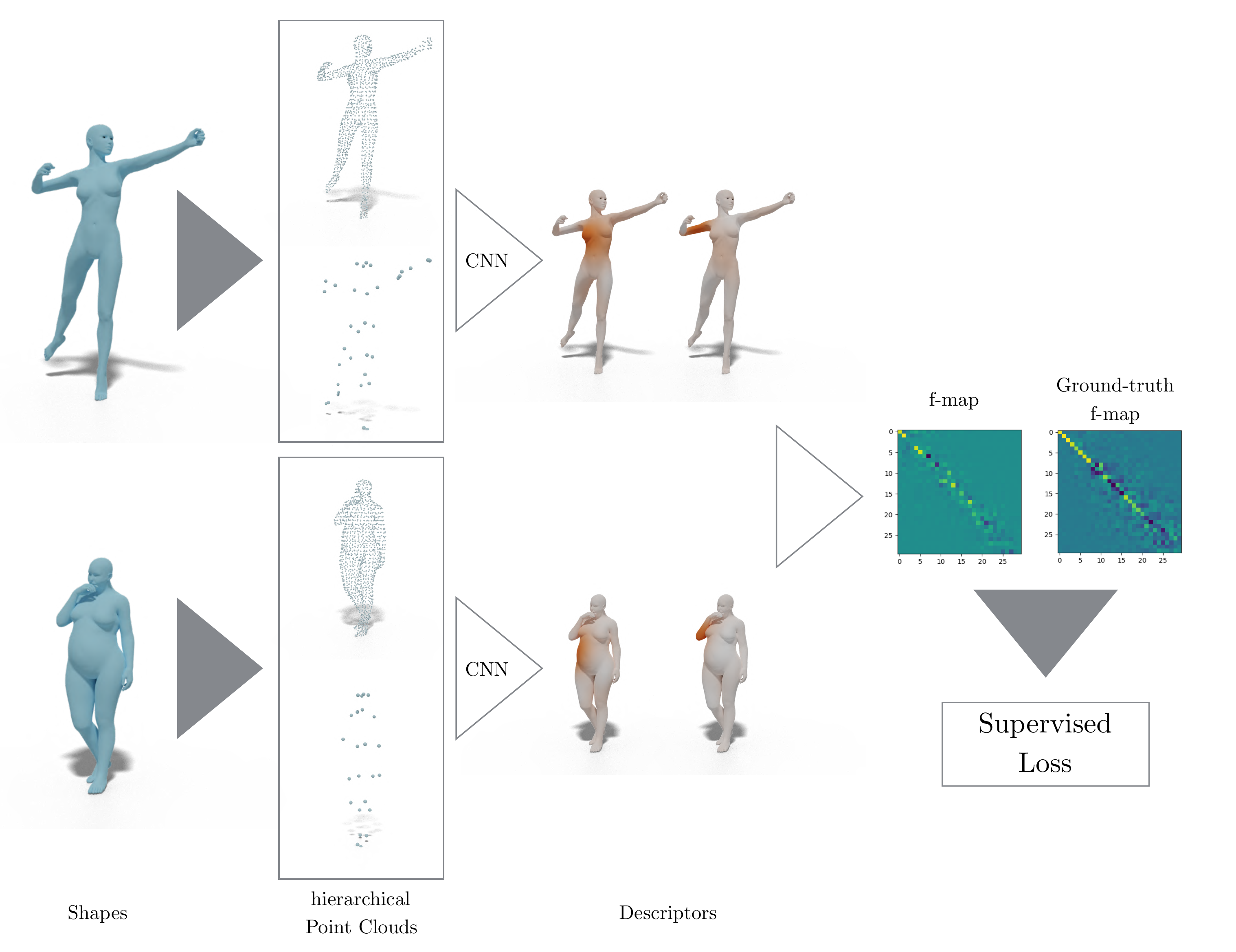}
\end{center}
   \caption{Pipeline of our method: 1) Down-sample source and target shapes with grid sampling (providing the pooling at different scales). 2) Learning point cloud characterizations and project them in the Laplace-Beltrami eigen basis. 3) Compute the functional map from source and target spectral descriptors, with our regularized fmap layer. 4) Compute the loss by comparing the computed functional map with the ground truth map.\label{fig:resume_method}}
\end{figure*}

It is remarkable that the descriptors learned on a parametric dataset such as the one used in 3D-CODED still generalize well to shapes with entirely different mesh connectivity and number of points. This is made possible by two components of our method. Firstly, it down-samples the input shapes through grid sampling before building these descriptor functions with convolutional neural networks. This allows for regularity in the input point clouds at all different hierarchies (see Figure \ref{fig:resume_method} for an example of such grid sampling). Secondly, the spectral projections take these point cloud descriptions to the shapes intrinsic space, adding some comprehensive surface-related information \textit{without depending too much on the connectivity}, like with SHOT descriptors. Without this intrinsic translation, the network could have trouble differentiating two geometric components close in euclidean space, such as for instance the arms in mesh $25$ of Figure \ref{fig:descs}.

Additionally, these descriptors seem to capture some segmentation information, such as for instance head, arms, body and legs for humans, as can be observed in Figure \ref{fig:descs}. More precise or complex descriptors such as hand or facial descriptors can not appear with only 30 eigen vectors. It is due to the fact that a spectral reconstruction of a human shape with only 30 eigenvectors does not show small details such as hands, feet or facial features.
One would need to push the number of eigenvectors above 100 to see such descriptors appear, and used correctly by the algorithm to produce even better correspondences. However, this could also more easily lead to overfitting.

\subsection{Additional Texture transfer on SHREC'19 re-meshed}

In Figure \ref{fig:shrec_maps} are shown additional qualitative results of our method (with and without Zoomout refinement \cite{melzi2019zoomout}), 3D coded \cite{groueix20183d}, FMNet \cite{litany2017deep} and Unsupervised FMNet \cite{halimi2018self} (with and without PMF refinement \cite{vestnerefficient}), trained on respectively 2000 and 100 shapes, as presented in the Results section, Experiment 2, of the main manuscript.

These results show again the failure of FMNet, due to the change in connectivity. It can be seen more thoroughly in the quantitative graphs provided in Figure \ref{fig:pr_graphs_2}.

This pair of shapes in Figure \ref{fig:shrec_maps} represents a challenging case for both 3D-CODED and our method. Indeed, \textit{these networks are not rotation invariant}, as discussed in the implementation section of the main manuscript. Here, the source shape is bent over and its head is really low compared with the rest of the body. 3D-CODED and our method are made robust to rotation around the vertical axis through data augmentation, but here the source shape is slightly rotated along another axis. As we can see, this resulted in poor reconstructions in the case of 3D-CODED algorithm, even with 2000 training shapes, whereas our method was able to yield good results with both a high and a low number of shapes.

\subsection{General Pipeline}
We also provide a visual illustration of our general pipeline in Figure \ref{fig:resume_method} to complement the textual description of our method provided in the main manuscript.

